%% file: DMLR-FORMAT.tex
\documentclass[twoside,11pt]{article}
\usepackage[preprint]{dmlr2e}

\usepackage{blindtext}

\usepackage[utf8]{inputenc} 
\usepackage[T1]{fontenc}    
\usepackage{url}            
\usepackage{amsfonts}       
\usepackage{nicefrac}       
\usepackage{microtype}      
\usepackage{lipsum}
\usepackage{fancyhdr}       
\usepackage{booktabs}
\usepackage{multirow}
\usepackage[table,xcdraw]{xcolor}
\usepackage{makecell}
\usepackage[inline]{enumitem}
\providecommand{\openreview}{}
\usepackage{lastpage}

\def\openreview{Pending}

\ShortHeadings{SEEC: External-Eye Dataset and Benchmark}{Nahass et. al. 2026}
\firstpageno{1}

\begin{document}

\title{SEEC: A Large-Scale Dataset and Benchmark for Self-Supervised External-Eye Representation Learning}

\author{
\name George R. Nahass$^{1,2,3}$ \email gnahas2@uic.edu \\
\name Michelle Fleishaker$^{1}$ \\
\name Sanjay Ganesh$^{1}$ \\
\name Sasha Hubschman$^{1}$ \\
\name R. V. Paul Chan$^{1,3}$ \\
\name Chad A. Purnell$^{4}$\\
\name Pete Setabutr$^{1,3}$ \\
\name Ann Q. Tran$^{1}$ \\
\name Sathya N. Ravi$^{2}$ \email sathya@uic.edu \\[0.5em]
\addr $^{1}$Department of Ophthalmology and Visual Sciences, University of Illinois Chicago, Chicago, IL, USA\\
$^{2}$Department of Computer Science, University of Illinois Chicago, Chicago, IL, USA\\
$^{3}$Artificial Intelligence in Ophthalmology, University of Illinois Chicago, Chicago, IL, USA \\
$^{4}$Department of Plastic Surgery, University of Illinois Chicago, Chicago, IL, USA
}

\editor{My editor}

\maketitle

\begin{abstract}
External-eye photography captures clinically relevant periocular anatomy at low acquisition cost, yet no large-scale standardized resource exists for evaluating self-supervised external-eye representations. We introduce a 4.35-million-image Standardized External-Eye Corpus (SEEC) derived from fourteen heterogeneous public face datasets, together with a benchmark for external-eye representation learning. For evaluation, we provide held-out evaluation distributions, landmark-annotated anatomical probe sets, and clinical disease-classification experiments spanning ptosis, thyroid eye disease, and craniofacial syndromes. We evaluate representations at two complementary levels: (a) the representation level, by assessing embedding geometry across multiple data distributions, and (b) the functional level, by assessing anatomical recovery, out-of-distribution anatomical transfer, and disease classification. Our models include ResNet-101 and ViT-B encoders pretrained using InfoNCE, VICReg, and LeJEPA objectives and compared with DINOv2 and MAE. Our results indicate that representation-level and functional-level evaluations are dissociable: broader embedding-space utilization does not consistently correspond to anatomical transfer or disease-classification performance, and no single encoder is universally optimal. Domain-specific external-eye encoders outperform DINOv2 and MAE on cross-dataset anatomical transfer despite being pretrained at substantially smaller scale, while external-eye-pretrained ViT models match or outperform these general-purpose baselines on frozen disease-classification probes. We release benchmark code, evaluation scripts, and preprocessing pipelines to reproduce SEEC at \href{https://monkeygobah.github.io/seec-project/}{the SEEC project page}.
\end{abstract}



\input{merged-sections/1-intro}
\input{merged-sections/1.5-related-work}

\input{merged-sections/2-construction}

\input{merged-sections/3-benchmark-resources}

\input{merged-sections/4-benchmark-tasks}
\input{merged-sections/5-results}
\input{merged-sections/7-conclusions}
\input{merged-sections/6-limitations}

\impact{
SEEC is intended to support development and evaluation of general-purpose external-eye representations using inexpensive visible-light imaging. More capable and reusable representations of external-eye anatomy could facilitate research on accessible ophthalmic screening, phenotyping, and oculomics, particularly in settings where specialized imaging equipment is unavailable. We also hope that the standardized corpus and benchmark will improve reproducibility by providing common data-processing and evaluation procedures for tasks in external-eye representation learning.

These potential benefits are accompanied by important risks. Periocular images remain potentially identifiable, and models trained on public face datasets may inherit demographic, acquisition, or sampling biases that are not fully characterized in SEEC. Representations learned from these data could also be repurposed for unintended applications such as identity recognition or used clinically without adequate external validation, potentially producing unreliable or inequitable results. As such, SEEC is not intended for face recognition, identity verification, or direct clinical deployment without appropriate external validation and oversight.}


\acks{This study was supported by an unrestricted grant from Research to Prevent Blindness, a donation from the Cless Family Foundation, Panorama Research Institute, Google Cloud, and the National Institutes of Health P30 EY001792 core grant. PS, GRN, and SR disclose equity holdings in Tagense Labs LLC. PS additionally discloses consulting relationships with Viatris and equity holdings in Lodestone. AQT discloses consulting and/or research relationships with Genentech/Roche, Argenx, Axogen, and Bausch + Lomb. SH discloses stock options in Horizon Therapeutics. CAP is on the medical advisory board of ImmersiveTouch Inc.}

\newpage
\bibliography{master-bib}

\newpage
\appendix
\input{merged-sections/99-release-and-maintenance}
\input{merged-sections/99-datasheet}
\input{merged-sections/99-supp-methods}

\input{merged-sections/99-supp-results}


\end{document}

%% file: merged-sections/1-intro.tex
\section{Introduction}

Unlike internal ocular imaging modalities such as fundus photography or optical coherence tomography, external-eye photography captures the visible anatomy of the eyelids, palpebral fissures, canthi, iris, sclera, brows, and surrounding periocular soft tissue. These structures are clinically relevant for documenting and assessing conditions such as ptosis, thyroid eye disease, facial nerve palsy, craniofacial anomalies, trauma, and postoperative outcomes \citep{ural2016utility, barman2025evaluation, kumar2024comprehensive,theng2021periorbital}.  More broadly, recent work suggests that external-eye photographs may encode biomarkers of systemic health with relevance in diabetes, kidney, and liver disease \citep{zhu2025oculomics,babenko2023deep}. The externally visible eye therefore provides a low-cost, camera-accessible anatomical interface that may support scalable biomarker discovery and systemic health monitoring \citep{babenko2023deep}.

Self-supervised learning has enabled transferable visual representations to be learned from large unlabeled image collections, and ophthalmic foundation-model development has benefited from large modality-specific datasets in retinal fundus photography, optical coherence tomography, and related device-based imaging \citep{zhou2023foundation,chia2024foundation}. Comparable large-scale resources for external-eye representation learning remain limited. Existing face datasets were developed for identity recognition, facial attributes, or image generation rather than standardized modeling of external-eye anatomy \citep{liu2015faceattributes,cao2018vggface2,karras2019style}. Dedicated iris and periocular datasets have generally been designed for biometric recognition or specialized eye-analysis tasks and are considerably smaller than contemporary pretraining corpora \citep{UBRISV1,UBRISV2,padole2012periocular,zanlorensi2022new,VISOB2,tiong2019periocular}. As a result, no existing resource provides the combination of scale, standardized external-eye anatomy, and evaluation structure needed to study general-purpose external-eye representations. Addressing this problem requires both a large, standardized pretraining corpus and a benchmark capable of assessing the quality, robustness, and transferability of external-eye representations across diverse datasets and downstream tasks.

Evaluating these representations therefore requires moving beyond any single downstream endpoint. Accordingly, here we consider the following three complementary questions:
\begin{enumerate*}[label=\textbf{\arabic*.}]
    \item how do learned representations occupy the embedding space across in-distribution and domain-shifted image distributions?
    \item do frozen representations preserve fine-grained periocular anatomy, and does this anatomical information transfer across datasets?
    \item do the learned representations support clinically relevant disease classification?
\end{enumerate*}
To this end, we evaluate representations at two complementary levels. At the representation level, we characterize embedding geometry across held-out and domain-shifted evaluation distributions. At the functional level, we evaluate anatomical landmark recovery, cross-dataset anatomical transfer, and clinical disease classification.

\textbf{Our contributions are threefold.} \textbf{\textit{First}}, we introduce a large-scale \textbf{Standardized External-Eye Corpus} (\textbf{SEEC}) containing 4.35 million aligned unilateral periocular crops derived from fourteen public facial image sources, with standardized preprocessing and fixed pretraining and evaluation subsets. \textbf{\textit{Second}}, we define a multi-axis benchmark designed to test whether external-eye representations preserve useful geometry and anatomy, transfer across datasets, and allow for clinically relevant downstream prediction. \textbf{\textit{Third}}, we establish reference baselines across contrastive, redundancy-reduction, and JEPA-style self-supervised objectives using ResNet-101 and ViT-B/16 backbones, with comparisons to DINOv2 and MAE. We additionally release benchmark code, preprocessing and reconstruction pipelines, train/evaluation splits, model training configurations, and evaluation scripts; processed images are released directly where source terms permit redistribution.


\begin{figure}[!t]
    \centering
    \includegraphics[width=1\linewidth]{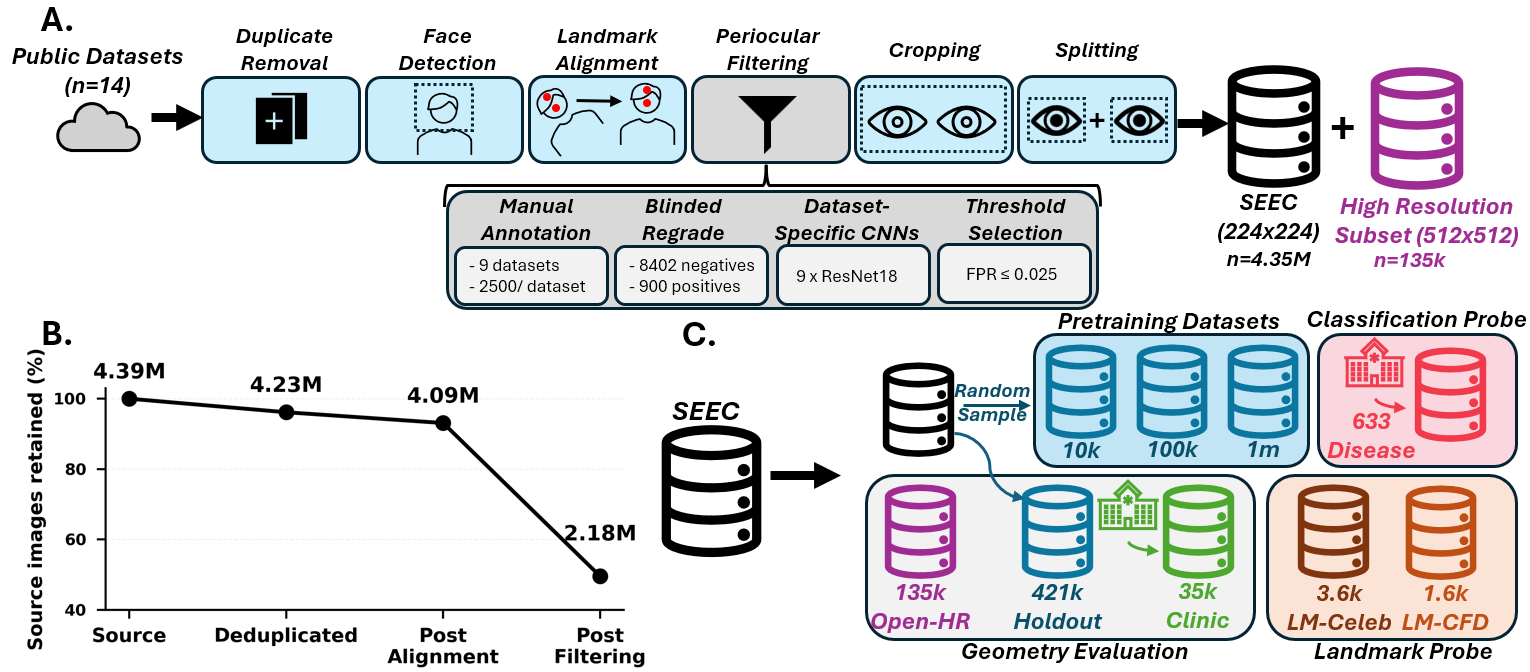}
    \caption{Overview of SEEC construction and benchmark resources.
    (A) SEEC construction pipeline. Fourteen public facial-image datasets were sequentially standardized and curated to produce 4.35 million unilateral $224\times224$ periocular images and a 135,000-image high-resolution subset. For datasets requiring automated periocular filtering, manual annotation and blinded regrading were used to train dataset-specific classifiers with source-specific filtering thresholds.
    (B) Source-image retention across major stages of SEEC construction. Counts through periocular filtering represent bilateral source images prior to left/right splitting.
    (C) Benchmark resources used for self-supervised pretraining and evaluation. Fixed SEEC subsets of 10,000, 100,000, and 1 million images were used for pretraining; held-out open-source and clinical datasets were used for embedding-geometry, anatomical-landmark, and disease-classification evaluations.}
    \label{fig:overview}
\end{figure}

a

%% file: merged-sections/1.5-related-work.tex
\section{Related Work}

\textbf{Deep learning and the external eye.}
External-eye images have been used for a growing range of clinical computer-vision tasks in oculoplastics, ophthalmology, and craniofacial medicine. Prior work includes blepharoptosis detection from periocular photographs \citep{hung2021deep,shu2024artificial,abascal2023development}, thyroid eye disease assessment using external photographs or orbital imaging \citep{karlin2023ensemble,moon2022machine,alkhadrawi2024deep,han2025deep}, and eyelid-lesion localization and classification \citep{li2022artificial,lee2023differentiating,hui2025deep}.

\textbf{Oculomics and ophthalmic representation learning.}
Oculomics studies ocular and periocular phenotypes as biomarkers of systemic, neurologic, developmental, and environmental health \citep{zhu2025oculomics}. Much of this literature has focused on internal ocular imaging, particularly retinal fundus photography and optical coherence tomography \citep{poplin2018prediction,ting2018eyeing}. These modalities have also become important for ophthalmic foundation-model development; RETFound, for example, used self-supervised pretraining on 1.6 million retinal images and was adapted across multiple downstream disease tasks \citep{zhou2023foundation}. External-eye oculomics has likewise demonstrated clinically relevant signal. Babenko et al. developed supervised deep-learning systems to predict diabetic retinal disease, HbA1c, and multiple systemic biomarkers directly from external-eye photographs \citep{babenko2022detection,babenko2023deep}. These external-eye studies focused on end-to-end prediction of predefined clinical endpoints rather than large-scale representation learning intended for reuse across tasks.

\textbf{Ocular and periocular datasets.}
Dedicated ocular and periocular datasets such as UBIRIS.v2, UBIPr, VISOB, and UFPR-Periocular have primarily been designed for iris recognition, periocular biometrics, verification, and related recognition tasks \citep{UBRISV1,UBRISV2,padole2012periocular,zanlorensi2022new,VISOB2,tiong2019periocular}. Other eye-image resources target specialized acquisition settings. For example, TEyeD contains millions of images from head-mounted eye trackers for gaze estimation, segmentation, and geometric eye modeling \citep{TEYED}. Large facial datasets such as CelebA, VGGFace2, and FFHQ also contain substantial periocular content, although they were originally developed for full-face recognition, facial attributes, and generative modeling \citep{liu2015faceattributes,cao2018vggface2,karras2019style}.

%% file: merged-sections/2-construction.tex
\section{SEEC Construction}

SEEC was constructed from fourteen heterogeneous publicly available facial image datasets containing visible ocular and periocular anatomy \cite{ma2015chicago, ma2021chicago, lakshmi2021india, de_Oliveira, Goldstein_2017,zhifei2017cvpr, venkatesh2020influence, Rothe-IJCV-2018, liu2015faceattributes, bansal2017umdfaces, karras2019style, cao2018vggface2}. The construction pipeline consisted of sequential preprocessing and curation stages including image deduplication, face detection and landmark-based alignment, manual periocular annotation, dataset-specific CNN-based filtering, bilateral periocular cropping, and division into unilateral eye crops. All retained unilateral images were standardized to $224\times224$ pixels to form the primary SEEC corpus, with a higher-resolution subset additionally standardized to $512\times512$ pixels. Figure~\ref{fig:overview}A summarizes the construction pipeline.

\subsection{Source Dataset Compilation}

Images were aggregated from fourteen publicly available facial image datasets: CFD, CFD-I, CFD-MR, Tufts, FEI, FIML, UTKFace, MORPH, WikiFace, CelebA, UMD Faces, IMDB-WIKI, FFHQ, and VGGFace2 \cite{ma2015chicago, ma2021chicago, lakshmi2021india, de_Oliveira, Goldstein_2017,zhifei2017cvpr, venkatesh2020influence, Rothe-IJCV-2018, liu2015faceattributes, bansal2017umdfaces, karras2019style, cao2018vggface2}. These sources span controlled facial-image datasets, celebrity image collections, and large-scale face-recognition datasets and provide variation in pose, illumination, image quality, resolution, acquisition setting, age, sex, and race/ethnicity. All datasets were obtained and processed in accordance with their source licensing restrictions (Supplementary Table \ref{tab:supp-license}). No modifications were made to the source images before preprocessing.

\subsection{Image Deduplication and Face Alignment}

An MD5 hash was computed for each image file within each source dataset to identify and remove exact duplicate images. For all remaining images, facial landmarks were detected using the MediaPipe FaceMesh model \cite{kartynnik2019real}. Images in which no face or multiple faces were detected were excluded from further processing. For images containing a single detected face, facial landmarks were used to standardize image orientation. Images were rotated such that the line connecting the soft-tissue nasion and midpoint of the hairline was vertically aligned, corresponding to MediaPipe landmark indices 6 and 10. The required rotation angle was computed from these landmarks and applied to the full image prior to subsequent periocular-region extraction and filtering.

\subsection{Periocular Image Annotation}

For each of the nine datasets requiring automated periocular filtering (CelebA, FIML, IMDB-WIKI, MORPH, UMD Faces, UTKFace, WikiFace, FFHQ, and VGGFace2), 2,500 images were randomly sampled and manually labeled for periocular visibility. Images were classified as periocular (positive) when both of the following criteria were satisfied: (1) both palpebral fissures were visible as distinct eyelid apertures, and (2) at least three of the four medial and lateral canthi were identifiable. Images were classified as non-periocular (negative) if either criterion was not satisfied. Negative examples therefore included images with periocular occlusion from hair, hands, or other objects; extreme head pose resulting in only one visible eye; image cropping that excluded part of the periocular region; or image degradation such as blur or compression that prevented reliable identification of eyelid structures.

To assess annotation consistency and refine the negative class, a second blinded grader independently reviewed randomly sampled subsets of images initially labeled as periocular ($n=100$) and non-periocular ($n=100$) from each dataset. In addition, all images initially classified as non-periocular were re-evaluated by the second grader. For classifier development, images classified as periocular by the initial grader were treated as positive, while images classified as non-periocular by both graders were treated as negative. Initially negative images for which the graders disagreed were designated ambiguous (gray). Ambiguous images were excluded from classifier training.

\subsection{CNN-Based Periocular Filtering}

For datasets in which periocular visibility could not be determined reliably from source structure alone, we trained dataset-specific classifiers and applied source-specific operating thresholds to retain images with adequate periocular visibility.

\subsubsection{Dataset-Specific CNN Training}

A separate ResNet-18 convolutional neural network was trained for each of the nine datasets requiring automated periocular filtering. For Tufts and FEI, frontal facial views could be identified using dataset-specific file-naming conventions. Images from CFD, CFD-I, and CFD-MR were sufficiently standardized such that all images remaining after the preceding preprocessing steps satisfied the predefined periocular visibility criteria and were retained. Each ResNet-18 was initialized with ImageNet-pretrained weights, and the final fully connected layer was replaced with a two-class output corresponding to periocular and non-periocular images. Training used the positive and negative annotations defined above.

For each dataset, the larger class was randomly subsampled to match the size of the smaller class. The resulting binary classification dataset was divided using stratified sampling into training, validation, and test sets, with 15\% assigned to validation and 15\% to testing. Input images were resized to 256 pixels and center-cropped to $224\times224$ pixels. Random horizontal flipping was used during training, and images were normalized using ImageNet mean and standard deviation values. Models were trained for 10 epochs using AdamW with a learning rate of $1\times10^{-4}$, weight decay of $1\times10^{-4}$, cosine learning-rate decay, and batch size 32. Model selection was based on validation F1 score.

\subsubsection{Dataset Filtering and Threshold Selection}

The trained dataset-specific classifiers were applied to all corresponding aligned images. For each dataset, thresholds between 0.50 and 0.95 were evaluated on the validation set. The final threshold was defined as the lowest threshold yielding a false-positive rate (FPR) of $\leq0.025$. If no evaluated threshold achieved this criterion, the highest threshold was selected to minimize retention of non-periocular images. Threshold selection was additionally assessed qualitatively by examining predicted probabilities for positive, negative, and ambiguous images and visually reviewing the 25 lowest-confidence images retained as positive at the selected threshold.

\subsection{Periocular Cropping and Standardization}

Following CNN-based filtering, retained facial images were cropped to the bilateral periocular region containing the eyes and brows. For each image, the original source image was first rotated using the landmark-derived alignment angle described above, with bicubic interpolation and canvas expansion to preserve the complete rotated image. Bilateral periocular bounding-box coordinates were then defined using MediaPipe Face Mesh landmarks 139 and 383 as the horizontal boundaries and landmarks 104 and 345 as the vertical boundaries, with the resulting box applied to the aligned image to generate the periocular crop.

Each bilateral periocular crop was subsequently divided into independent right- and left-eye images along a stored vertical split coordinate; when no explicit split coordinate was available, the horizontal midpoint of the crop was used. The left portion of the image was assigned to the subject's right eye (OD), and the right portion was assigned to the subject's left eye (OS), consistent with the orientation of frontal facial photographs.

The resulting unilateral periocular images were standardized to fixed spatial resolutions. All images were resized to $224\times224$ pixels using Lanczos resampling to form the primary SEEC corpus for self-supervised pretraining. A higher-resolution subset was additionally generated at $512\times512$ pixels using Lanczos resampling. To avoid excessive upsampling, images were included in the high-resolution subset only when the original unilateral periocular crop had a minimum side length of at least 96 pixels before resizing.

%% file: merged-sections/3-benchmark-resources.tex

\section{Benchmark Resources}

From the primary SEEC corpus, we define fixed pretraining subsets and held-out open-source evaluation distributions for representation-learning experiments (Table~\ref{tab:datasets}; Figure~\ref{fig:overview}C). We additionally use open-source landmark-annotated periocular datasets for anatomical probing and clinical image repositories for domain-shift and disease-classification evaluation.

\begin{table}[]
\centering
\begin{tabular}{@{}ccccc@{}}
\toprule
\textbf{Dataset}    & \textbf{Role}                         & \textbf{N} & \textbf{Labels}                 & \textbf{Public Release}   \\ \midrule
SEEC & Master Corpus                           & 4,352,692  & Unlabeled                       & Partial†                  \\ \midrule
Pretrain-10K        & \multirow{3}{*}{Pretraining}          & 10,000     & \multirow{3}{*}{Unlabeled}      & \multirow{3}{*}{Partial†} \\
Pretrain-100K &                   & 100,000   &                &       \\
Pretrain-1M   &                   & 1,000,000 &                &       \\ \midrule
Holdout             & \multirow{2}{*}{Open-Source Geometry} & 421,730    & \multirow{3}{*}{Unlabeled}      & \multirow{2}{*}{Partial†} \\
Open-HR       &                   & 134,969   &                &       \\
Clinic        & Clinical Geometry & 35,082    &                & None‡ \\ \midrule
LM-Celeb            & \multirow{2}{*}{Landmark Probe}       & 3,610      & \multirow{2}{*}{Nine Landmarks} & \multirow{2}{*}{Partial†} \\
LM-CFD        &                   & 1,596     &                &       \\ \midrule
Disease       & Disease Classification        & 633       & Disease Labels & None‡ \\ \bottomrule
\end{tabular}%
\caption{Dataset components and their roles in the benchmark. $^\dagger$ indicates components released directly where source licenses permit redistribution of derived images; for the remaining public-source components, we provide preprocessing code, manifests, crop coordinates, and reconstruction tools to allow authorized users to rebuild the benchmark. $^\ddagger$ indicates clinical components used under institutional data-use restrictions and not publicly released.}
\label{tab:datasets}
\end{table}

\subsection{Pretraining and Open-Source Evaluation Splits}

From the primary $224\times224$ SEEC corpus, we define fixed pretraining subsets and held-out open-source evaluation distributions (Table~\ref{tab:datasets}; Figure~\ref{fig:overview}C).

\textbf{Pretraining subsets.}
Fixed subsets of 10,000, 100,000, and 1,000,000 images were sampled from SEEC for controlled self-supervised pretraining experiments evaluating the effect of unlabeled data scale. These subsets are referred to as \textit{Pretrain-10K}, \textit{Pretrain-100K}, and \textit{Pretrain-1M}.

\textbf{In-distribution evaluation.}
A held-out set of 421,730 images was sampled from the same source datasets and preprocessing pipeline as the pretraining corpus but excluded from self-supervised training. This set provides an in-distribution evaluation resource without direct image overlap and is referred to as \textit{Holdout}.

\textbf{High-resolution open-source evaluation.}
The 134,969-image high-resolution SEEC subset defined during corpus construction was reserved as an additional held-out evaluation distribution. These images were excluded from self-supervised pretraining and resized to $224\times224$ pixels for evaluation. This set is referred to as \textit{Open-HR}.

\subsection{Anatomical Landmark Probe Datasets}

For anatomical evaluation, we constructed landmark probe datasets from public CelebA- and CFD-derived periocular segmentation resources by pairing each source image with its semantic mask, splitting the face at the midline into unilateral eye crops, and resizing each crop to 224$\times$224 pixels \cite{nahass2025open}. Landmark coordinates were derived algorithmically from the available segmentation masks using OpenCV tools with the resized image, consistent with prior work deriving periocular landmarks from anatomically annotated segmentation boundaries \cite{van2021periorbitai}. Because the underlying segmentation masks were annotated according to anatomically defined periocular structures, this method allows for a reproducible mapping from semantic boundaries to landmark coordinates while minimizing further annotation burden. Briefly, an inscribed circle was fit to the iris mask, its center defined the iris centroid, and its superior and inferior points defined the corresponding iris landmarks. Scleral and eyelid superior/inferior landmarks were defined by mask maximum and minimum along the vertical axis passing through the iris centroid, while medial and lateral canthi were defined from the horizontal extrema of the visible scleral region. This procedure yielded nine landmarks per crop: iris centroid, iris superior, iris inferior, sclera superior, sclera inferior, medial canthus, lateral canthus, lid superior, and lid inferior. The resulting landmark probe datasets are referred to as \textit{LM-Celeb} and \textit{LM-CFD}.

\subsection{Clinical Evaluation Datasets}

To evaluate clinical domain shift, we use a repository of \textit{unlabeled} 35,082 external-eye photographs collected over 20 years from a large craniofacial center. These images span diverse conditions affecting the eyes, eyelids, periocular soft tissue, and craniofacial anatomy. This dataset provides a clinically realistic distribution for evaluating whether pretrained representations remain well behaved outside curated public-image settings. This is referred to as \textit{Clinic}.

For disease classification, we constructed a labeled clinical dataset containing 633 diseased-eye images across five disease classes: thyroid eye disease, chalazion, Treacher Collins syndrome, ptosis, and Crouzon/Apert/Pfeiffer syndromes. The 633 images represent a separate dataset collected at a different clinic for which labels were available rather than a deliberate subsample of \textit{Clinic}. Disease classes were selected because they produce visible periocular or craniofacial findings and are clinically relevant to external-eye assessment. All images were reviewed by an oculoplastic surgery fellow. Images were resized to $224\times224$ pixels prior to evaluation. This dataset is used only for downstream five-class disease-classification experiments and is referred to as \textit{Disease}.

Both clinical datasets were captured during routine clinical care across a wide range of devices and acquisition conditions, resulting in natural variation in lighting, pose, framing, resolution, and background. IRB approval was obtained before use.

\subsection{Release, Licensing, and Intended Use}

Because SEEC is derived from multiple source datasets with different licensing and use terms, release is organized according to source-specific redistribution provisions (Table~\ref{tab:supp-license}). We will release processed periocular crops for sources whose terms permit redistribution of derived images. For sources for which redistribution is restricted or not expressly permitted, we provide end-to-end tools that allow authorized users to reconstruct the corresponding SEEC resources from the original source datasets.

Access to clinical evaluation data may be considered only under appropriate institutional approval, data-use agreements, and ethical review. SEEC is not intended for face recognition or identity verification. Standardized periocular crops reduce full-face exposure but do not completely eliminate privacy or identifiability concerns; source restrictions and intended uses will therefore be documented as part of the release. Code, training pipelines, and evaluation protocols are available
\href{https://github.com/monkeygobah/seec-benchmark}{\texttt{here}}.
All scripts required to reconstruct SEEC can be found
\href{hhttps://github.com/monkeygobah/seec-dataset}{\texttt{here}}.

%% file: merged-sections/4-benchmark-tasks.tex
\begin{figure}[!t]
    \centering
    \includegraphics[width=1\linewidth]{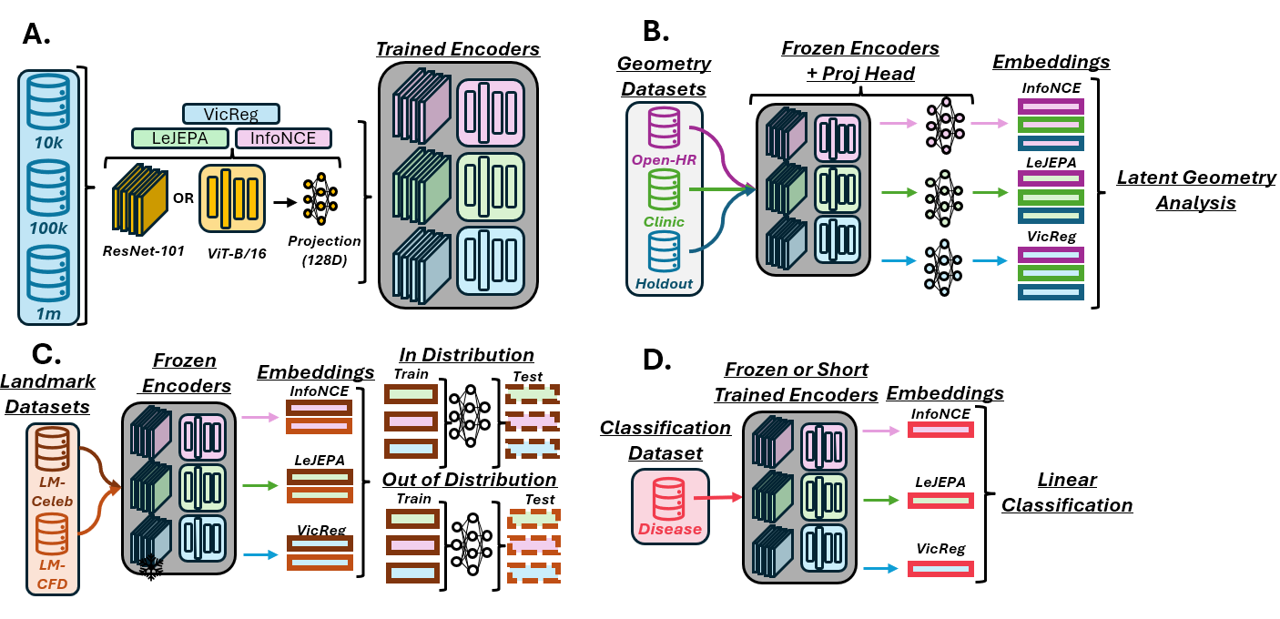}
    \caption{Overview of the benchmark tasks. (A) Encoders are pretrained under a controlled self-supervised benchmark strategy. (B) Frozen embeddings are evaluated using geometry metrics across distributions. (C) Frozen embeddings are used for anatomical landmark probing in in-distribution and cross-dataset transfer settings. (D) Disease classification is evaluated using frozen linear probes and five-epoch fine-tuning. For embeddings, border color denotes the evaluation dataset, and fill color denotes the pretraining objective or baseline model.}
    \label{fig:benchmark-desc}
\end{figure}

\section{Benchmark Task Descriptions}

We define four benchmark tasks for evaluating external-eye representations: embedding geometry, in-distribution anatomical landmark recovery, out-of-distribution anatomical transfer, and disease classification. Figure~\ref{fig:benchmark-desc} summarizes the benchmark design.

\subsection{Model Setup and Regularizers}\label{model-setup}

We evaluate external-eye self-supervised pretraining using various architectures, objective families, initializations, and pretraining data scales. Models are pretrained on standardized $224\times224$ unilateral external-eye crops from \textit{Pretrain-10K}, \textit{Pretrain-100K}, or \textit{Pretrain-1M}. Unless otherwise noted, downstream evaluations are performed using frozen backbone encoders. For both architectures, we evaluate random initialization and ImageNet initialization in the geometry benchmark. 

We evaluate two backbone architectures, ResNet-101 and ViT-B/16, under the same 50{,}000-step training budget. Across architectures, self-supervised training uses a three-layer projection head with hidden dimension 2048 and output dimension 128. The backbone feature dimension is architecture-specific. We compare three representative self-supervised objective families: contrastive learning, redundancy-reduction learning, and JEPA-style predictive learning, instantiated as InfoNCE, VICReg, and LeJEPA, respectively \cite{oord2018representation,bardes2021vicreg,balestriero2025lejepa}. 

\subsubsection{Evaluated Objectives}

To efficiently utilize the representation space, we include regularization functions that impose complementary constraints on paired augmented views and on the projected embedding space while pretraining. InfoNCE \cite{oord2018representation} is a contrastive objective that aligns embeddings from two views of the same image while contrasting them against other images in the minibatch using temperature-scaled cosine similarity. VICReg \cite{bardes2021vicreg} combines invariance, variance, and covariance regularization,
\[
\mathcal{L}_{\mathrm{VICReg}}
=
\lambda\mathcal{L}_{\mathrm{inv}}
+
\mu\mathcal{L}_{\mathrm{var}}
+
\nu\mathcal{L}_{\mathrm{cov}},
\]
where the invariance term aligns paired views, the variance term prevents collapse of individual embedding dimensions, and the covariance term penalizes redundancy across dimensions. LeJEPA \cite{balestriero2025lejepa} uses a JEPA-style predictive objective with Sketched Isotropic Gaussian Regularization (SigReg) to prevent representation collapse,
\[
\mathcal{L}_{\mathrm{LeJEPA}}
=
(1-\lambda)\mathcal{L}_{\mathrm{sim}}
+
\lambda\mathcal{L}_{\mathrm{SigReg}},
\]
where the similarity term encourages agreement between paired views and SigReg regularizes the embedding distribution. In our experiments, all objectives are trained using the same two-view augmentation protocol to allow for fair comparisons across methods. Please see Appendix~\ref{app:ssl_objectives} for implementation details of the objectives and training settings considered.

For comparison with general-purpose visual representations, we also evaluated frozen ViT representations from DINOv2 ViT-B pretrained on LVD-142M and MAE ViT-B pretrained on ImageNet-1K \cite{oquab2023dinov2,he2022masked}.

\subsection{Embedding Geometry Under Distribution Shift}

For each frozen model and evaluation distribution, we compute geometric summary statistics to characterize how the learned embedding space is occupied (Figure~\ref{fig:benchmark-desc}B). A well-utilized embedding space distributes variance broadly across dimensions rather than collapsing to a low-dimensional subspace, a property that self-supervised objectives explicitly or implicitly encourage \cite{chen2020simple,bardes2021vicreg}. 

We first compute geometry metrics on centered projected embeddings $\tilde{X}=X-\bar{X}$, where $X \in \mathbb{R}^{N \times D}$, and $\lambda_1 \geq \cdots \geq \lambda_D$ denote the eigenvalues of the embedding covariance matrix.

Our primary metric is the participation-ratio effective rank, $r_{\mathrm{eff}}={(\sum_{i=1}^{D}\lambda_i)^2}/{\sum_{i=1}^{D}\lambda_i^2}$, which we report after normalizing by the number of embedding dimensions as $r_{\mathrm{eff}}/D$. Higher values indicate broader use of the embedding dimensions, whereas lower values indicate concentration of variance in fewer directions. We evaluate $r_{\mathrm{eff}}/D$ across \textit{Holdout}, \textit{Open-HR}, and \textit{Clinic} and interpret it as a representation-level diagnostic rather than a direct measure of functional utility. Additional geometry summaries are reported in Appendix~\ref{app:embedding_geometry}.

\begin{figure}
    \centering
    \includegraphics[width=1\linewidth]{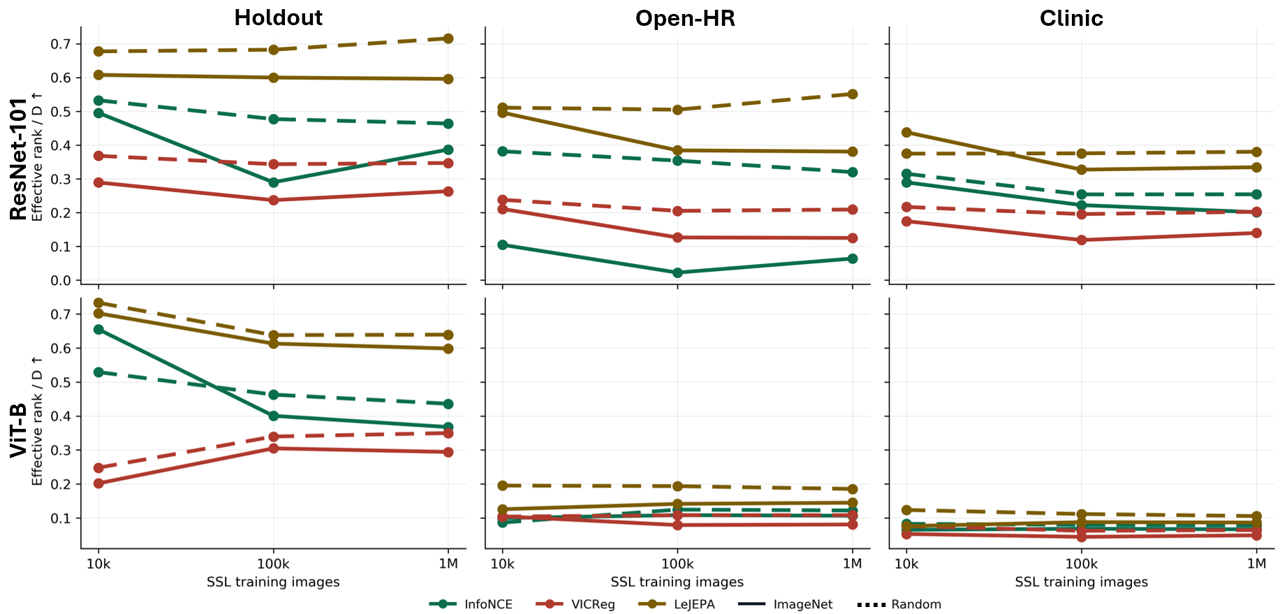}
    \caption{Effective rank normalized by feature dimension for ResNet-101 (top) and ViT-B (bottom) projector representations across training-set scale, SSL objective, initialization, and evaluation datasets.}
    \label{fig:erank}
\end{figure}

\subsection{Frozen Anatomical Landmark Probes}\label{sec:probes}

To evaluate whether frozen representations preserve anatomically meaningful periocular structure, we train lightweight landmark regression probes on top of extracted backbone embeddings (Figure~\ref{fig:benchmark-desc}C). Only models pretrained with \textit{Pretrain-1M} were used on this task. We evaluate landmark prediction on \textit{LM-Celeb} and \textit{LM-CFD}, each of which contains two-dimensional annotations for nine ophthalmic landmarks spanning iris, scleral, and eyelid anatomy. We consider both in-distribution and out-of-distribution probing. In-distribution probes are trained, validated, and tested on fixed splits from the same landmark dataset, while out-of distribution probes are trained on \textit{LM-Celeb} embeddings and evaluated on \textit{LM-CFD} embeddings without retraining.

Before probe training, $4\times4$ pooling is applied to frozen backbone feature maps. For ViT models, patch tokens are first reshaped into spatial feature maps. Additional probe architecture, optimization, and metric details are provided in Appendix~\ref{app:landmark_probe}.

\begin{figure}[!t]
    \centering
    \includegraphics[width=1\linewidth]{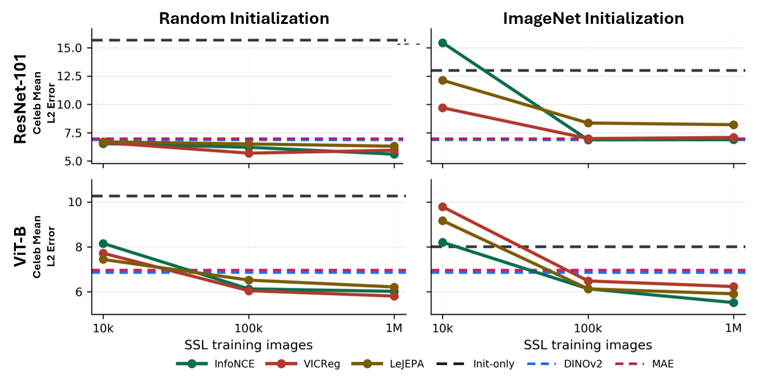}
    \caption{Within-dataset LM-Celeb landmark probe performance for ResNet-101 and ViT encoders. Lines show SSL objectives across training-set scale; dashed references show init-only and external ViT baselines.}
    \label{fig:lm-within}
\end{figure}

\subsection{Frozen and Fine-Tuned Disease Classification Probes}

To evaluate whether pretrained external-eye representations encode disease-relevant structure, we perform five-class disease classification on \textit{Disease} using ResNet-101 and ViT-B/16 encoders initialized randomly prior to self-supervised pretraining on \textit{Pretrain-1M}. The dataset was divided into an 80/20 train/test split using a fixed random seed and stratification by disease class. Splits were grouped at the image level, such that all crops derived from the same source image were assigned to the same partition in order to prevent leakage.

For each encoder, we first evaluate frozen representation quality using a linear probe. Encoder weights are fixed, embeddings are extracted after $4\times4$ pooling, and only a linear classifier is trained on the grouped training split. We additionally evaluate short-horizon supervised adaptation by jointly fine-tuning each encoder and linear classifier for five epochs on the same training split (Figure~\ref{fig:benchmark-desc}D). Frozen and fine-tuned models are evaluated on the same grouped held-out test split.

%% file: merged-sections/5-results.tex
\section{Results}

\begin{table}[]
\centering
 \resizebox{\columnwidth}{!}{%
\begin{tabular}{@{}c|ccccccc@{}}
\toprule
Model                       & Initialization            & Objective & Mean L2 & Mean MAE & Iris Centroid & Medial Canthus & Lateral Canthus \\ \midrule
\multirow{2}{*}{ViT} & DinoV2                  & DinoV2   & 5.04          & 3.23          & 4.91          & 5.55          & 5.67          \\
                     & MAE                     & MAE      & 8.89          & 5.88          & 7.53          & 9.80          & 10.47         \\ \midrule
\multirow{8}{*}{ResNet-101} & \multirow{4}{*}{ImageNet} & Baseline  & 12.08   & 7.90     & 10.95         & 15.04          & 13.36           \\
                     &                         & InfoNCE  & 6.04          & 3.87          & 5.40          & 6.55          & 6.67          \\
                     &                         & LeJEPA   & 6.82          & 4.36          & 6.17          & 8.81          & 7.47          \\
                     &                         & VicReg   & 5.51          & 3.51          & 5.13          & 7.00          & 6.01          \\ \cmidrule(l){2-8} 
                     & \multirow{4}{*}{Random} & Baseline & 14.24         & 8.39          & 7.38          & 39.21         & 34.33         \\
                     &                         & InfoNCE  & 10.33         & 6.95          & 9.27          & 11.28         & 10.80         \\
                     &                         & LeJEPA   & 5.16          & 3.29          & 5.05          & 6.87          & 6.46          \\
                     &                         & VicReg   & 4.94          & 3.16          & 4.55          & 5.61          & 5.59          \\ \midrule
\multirow{8}{*}{ViT}        & \multirow{4}{*}{ImageNet} & Baseline  & 6.57    & 4.21     & 5.85          & 7.28           & 7.15            \\
                     &                         & InfoNCE  & 5.13          & 3.29          & 5.01          & 6.31          & 5.61          \\
                     &                         & LeJEPA   & 5.29          & 3.35          & 5.84          & 6.63          & 5.49          \\
                     &                         & VicReg   & 4.79          & 3.04          & 4.40          & 5.89          & 4.96          \\ \cmidrule(l){2-8} 
                     & \multirow{4}{*}{Random} & Baseline & 7.21          & 4.55          & 6.53          & 8.24          & 9.19          \\
                     &                         & InfoNCE  & \textbf{4.24} & \textbf{2.65} & \textbf{3.40} & 6.16          & 4.84          \\
                     &                         & LeJEPA   & 4.42          & 2.83          & 4.23          & \textbf{5.51} & \textbf{4.67} \\
                     &                         & VicReg   & 4.60          & 2.92          & 4.22          & 5.82          & 5.55          \\ \bottomrule
\end{tabular}%
}
\caption{Celeb-to-CFD landmark transfer performance for frozen ResNet-101 and ViT encoders. Probes were trained and selected on LM-Celeb and evaluated on LM-CFD test images; errors are reported in 224-pixel crop coordinates.}
\label{tab:lm-transfer}
\end{table}

\subsection{SEEC Construction and Curation}

\subsubsection{Corpus Assembly and Manual Annotation}

Across the fourteen source datasets, 4,394,599 images were initially identified. Removal of 169,383 exact within-dataset duplicates yielded 4,225,216 images after deduplication. Following MediaPipe-based face detection, landmark alignment, and source-specific preprocessing, 4,090,709 images remained for periocular filtering. Source-level composition and attrition across construction stages are reported in Supplementary Tables~\ref{tab:supp-source-composition} and~\ref{tab:supp-final-composition}.

For the nine datasets requiring automated periocular filtering, 2,500 images per dataset were randomly sampled for manual annotation, yielding 22,500 unique annotated images. The initial grader classified 14,098 images as periocular and 8,402 as non-periocular. A second blinded grader subsequently re-evaluated all 8,402 images initially classified as non-periocular, as well as a random sample of 100 initially positive images from each dataset ($n=900$). Thus, the annotation procedure comprised 31,802 individual grading decisions across the two graders.

Intergrader agreement was consistently high for images initially classified as periocular, ranging from 0.95 to 1.00 across datasets. Agreement for the randomly sampled initially negative images ranged from 0.39 to 0.95, while agreement across the complete regrade of all initially negative images ranged from 0.28 to 0.95. Images remaining uncertain after regrading were assigned to the ambiguous (gray) category and excluded from classifier training. Dataset-specific annotation distributions and agreement statistics are reported in Supplementary Tables~\ref{tab:supp-grader-agreement} and~\ref{tab:supp-final-labels}.

\subsubsection{CNN-Based Periocular Filtering}

Dataset-specific ResNet-18 classifiers demonstrated strong discrimination between periocular and non-periocular images, with held-out test accuracy ranging from 0.85 to 1.00 and AUROC from 0.95 to 1.00 (Supplementary Table~\ref{tab:supp-cnn-perf-thresholds}). Rather than applying a common decision threshold across datasets, operating points were selected independently for each source to limit retention of non-periocular images. Selected thresholds ranged from 0.50 to 0.95.

At these operating points, held-out FPR ranged from 0.000 to 0.036 and TPR from 0.59 to 1.00. Seven of nine classifiers achieved the prespecified FPR target of $\leq0.025$; FIML and FFHQ reached FPRs of 0.028 and 0.036, respectively, at the most conservative evaluated operating point. The resulting calibration therefore prioritized limiting false retention of non-periocular images over maximizing sensitivity. For each dataset, the 25 lowest-confidence images retained at the selected threshold were additionally reviewed visually. Representative retained images and prediction-score distributions are shown in Supplementary Figure~\ref{fig:supp-filtering-qc}. 


\subsubsection{Final Corpus Composition}

The 2,176,346 retained bilateral periocular images were divided into independent right- and left-eye crops, yielding 4,352,692 unilateral periocular images standardized to $224\times224$ pixels (Figure \ref{fig:overview}). A higher-resolution subset of 134,969 unilateral images meeting the prespecified native-resolution criterion was additionally generated at $512\times512$ pixels. Source-specific counts across all major construction stages are reported in Supplementary Table~\ref{tab:supp-final-composition}, and the overall construction workflow is summarized in Figure~\ref{fig:overview}A.

\subsection{Analysis of Embedding Geometry}

We first analyzed the geometry of the learned projected representation spaces using effective rank normalized by embedding dimension ($r_\mathrm{eff}/D$) across \textit{Holdout}, \textit{Open-HR}, and \textit{Clinic} (Figure~\ref{fig:erank}). Across both ResNet-101 and ViT-B/16, LeJEPA produced the highest $r_\mathrm{eff}/D$ among the evaluated self-supervised objectives, indicating broader use of the available projected embedding dimensions. For ResNet-101, the strongest LeJEPA configurations reached $r_\mathrm{eff}/D=0.72$ on \textit{Holdout}, $0.55$ on \textit{Open-HR}, and $0.44$ on \textit{Clinic}. For ViT-B/16, LeJEPA also achieved the highest effective rank on each distribution, reaching $r_\mathrm{eff}/D=0.73$ on \textit{Holdout}, $0.20$ on \textit{Open-HR}, and $0.12$ on \textit{Clinic}.

Additionally, distribution shift substantially affected representation geometry. Although the relative ordering of objective families was broadly stable, the magnitude of separation varied across \textit{Holdout}, \textit{Open-HR}, and \textit{Clinic}. All objectives showed reduced $r_\mathrm{eff}/D$ under distribution shift, with the largest reductions on \textit{Clinic}. This effect was especially pronounced for ViT-B/16, where projected representations that were comparatively high-rank on \textit{Holdout} became much lower-rank on the clinically shifted distribution. These results indicate that geometry conclusions drawn from source-matched data do not transfer unchanged to external or clinical settings.

Initialization also affected embedding-space utilization. Randomly initialized models generally produced higher-rank projected embeddings than ImageNet-initialized models, particularly for InfoNCE and VICReg on \textit{Open-HR} and \textit{Clinic}. Together, these results show that embedding geometry is sensitive to objective family, initialization, architecture, and evaluation distribution. We therefore interpret $r_\mathrm{eff}/D$ as a representation-level diagnostic, while downstream anatomical and disease-relevant utility are evaluated separately in the following sections. Additional geometry metrics demonstrated consistent trends; further evaluation of these is provided in Appendix~\ref{app:embedding_geometry_results}.

\begin{figure}
    \centering
    \includegraphics[width=\linewidth]{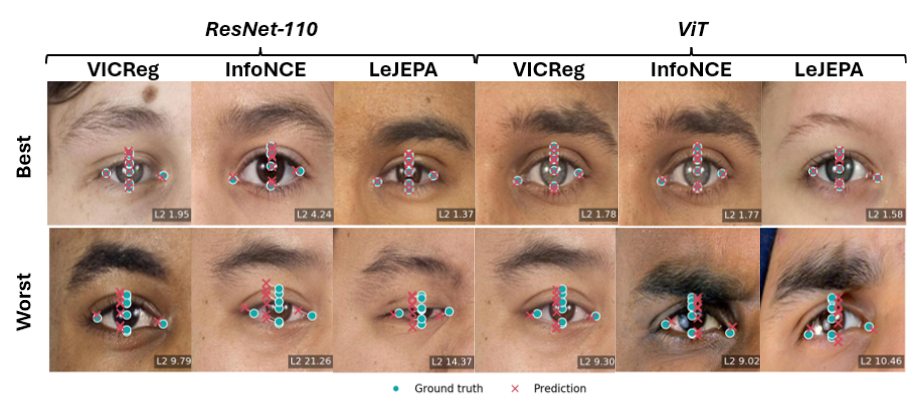}
    \caption{Representative Celeb-to-CFD landmark predictions from out-of-distribution landmark probes. Columns show probes trained on 1M-image random-initialized ResNet-101 and ViT models using VICReg, InfoNCE, or LeJEPA. For each objective, the top row shows the best CFD test example and the bottom row shows the worst example by mean landmark L2 error. Ground-truth landmarks are shown in teal and predictions in red.}
    \label{fig:lm-qual-transfer}
\end{figure}

\subsection{Anatomical Landmark Probes}

We next evaluated whether frozen representations preserve anatomically meaningful external-eye structure using landmark prediction probes on \textit{LM-Celeb} and \textit{LM-CFD}. We report both in-distribution performance, where probes were trained and evaluated on splits from the same landmark dataset, and out-of-distribution performance, where probes were trained on \textit{LM-Celeb} and evaluated on \textit{LM-CFD} (Figure~\ref{fig:benchmark-desc}C).

\subsubsection{In-Distribution}

Across both encoder families, self-supervised external-eye pretraining substantially improved landmark recovery relative to initialization-only baselines. This effect was observed for both \textit{LM-Celeb} (Figure~\ref{fig:lm-within}) and \textit{LM-CFD} (Figure~\ref{fig:cfd-within}), indicating that the learned representations preserve spatial information relevant for anatomical recovery on both datasets evaluated. Performance generally improved with increasing pretraining scale, although the best objective varied across dataset, architecture, and initialization. ResNet-101 and ViT-B encoders both reached low mean landmark error after pretraining, with ViT-B/16 models achieving slightly lower errors overall.

\subsubsection{Out-of-Distribution}

In the out-of-distribution task, among ResNet-101 encoders, random-initialized VICReg achieved the lowest overall error (mean L2 $=4.94$), followed by random-initialized LeJEPA (mean L2 $=5.16$). ViT-B/16 encoders performed better overall, with random-initialized InfoNCE achieving the lowest mean L2 and mean MAE across all evaluated encoders (mean L2 $=4.24$, mean MAE $=2.65$; Table~\ref{tab:lm-transfer}). Random-initialized ViT-B/16 LeJEPA achieved the lowest medial and lateral canthus errors.

Domain-specific external-eye pretraining was superior to the general-purpose ViT foundation-model baselines in the cross-dataset landmark transfer setting. The best external-eye-pretrained ViT encoders outperformed both DINOv2 and MAE on mean L2, mean MAE, iris centroid error, medial canthus error, and lateral canthus error. Thus, despite being trained at substantially smaller scale than DINOv2, domain-specific external-eye ViT-B/16 pretraining produced representations that were more useful for cross-dataset periocular landmark transfer. Figure~\ref{fig:lm-qual-transfer} provides qualitative examples of landmark predictions for representative out-of-distribution models.

\subsection{Disease Classification Probe}

We next evaluated whether pretrained representations support disease classification from external-eye images using frozen linear probing and short-horizon supervised adaptation (Figure~\ref{fig:benchmark-desc}D).

\subsubsection{Frozen Probe}

In the frozen linear-probe setting, external-eye-pretrained ViT-B/16 models performed strongest overall (Table~\ref{tab:classification}). ViT-B/16 InfoNCE achieved the highest accuracy, recall, and F1, while ViT-B/16 VICReg was the second-best model across all three metrics. DINOv2 was a strong general-purpose baseline, matching LeJEPA in accuracy and F1, but remained below the best external-eye-pretrained ViT-B/16 models. MAE and all ResNet-101 models performed lower overall. These results indicate that disease-relevant information is linearly recoverable from external-eye-pretrained ViT-B/16 representations and that transformer-scale external-eye pretraining improves frozen disease separability relative to the corresponding ResNet-101 encoders.

\begin{table}[]
\centering
\begin{tabular}{@{}c|ccccc@{}}
\toprule
\textbf{} & Model                       & Objective & Accuracy      & Recall        & F1            \\ \midrule
\multirow{8}{*}{\makecell{Training of \\ Encoder and Classifier \\ for 5 Epochs}} & \multirow{2}{*}{ViT-B/16} & DINOv2  & 0.40          & 0.26          & 0.34          \\
          &                             & MAE       & \textbf{0.89} & \textbf{0.85} & \textbf{0.89} \\ \cmidrule(l){2-6} 
          & \multirow{3}{*}{ResNet-101} & InfoNCE   & 0.71          & 0.66          & 0.70          \\
          &                             & LeJEPA    & 0.56          & 0.47          & 0.53          \\
          &                             & VICReg    & 0.71          & 0.66          & 0.70          \\ \cmidrule(l){2-6} 
                                                                 & \multirow{3}{*}{ViT-B/16} & InfoNCE & 0.87          & 0.82          & \textit{0.88} \\
          &                             & LeJEPA    & \textit{0.88} & 0.82          & \textit{0.88} \\
          &                             & VICReg    & \textbf{0.89} & \textit{0.83} & \textbf{0.89} \\ \midrule
\multirow{8}{*}{Frozen}                                          & \multirow{2}{*}{ViT-B/16} & DINOv2  & 0.88          & 0.82          & 0.88          \\
          &                             & MAE       & 0.84          & 0.78          & 0.84          \\ \cmidrule(l){2-6} 
          & \multirow{3}{*}{ResNet-101} & InfoNCE   & 0.83          & 0.75          & 0.82          \\
          &                             & LeJEPA    & 0.83          & 0.78          & 0.83          \\
          &                             & VICReg    & 0.83          & 0.74          & 0.82          \\ \cmidrule(l){2-6} 
                                                                 & \multirow{3}{*}{ViT-B/16} & InfoNCE & \textbf{0.91} & \textbf{0.87} & \textbf{0.91} \\
          &                             & LeJEPA    & 0.88          & 0.83          & 0.88          \\
          &                             & VICReg    & \textit{0.90} & \textit{0.85} & \textit{0.90} \\ \bottomrule
\end{tabular}%
\caption{Disease classification performance on held-out external-eye images using frozen linear probes and five-epoch joint training of the full backbone and linear classifier. Accuracy, F1, and recall are reported on the held-out test set; F1 and recall are averaged across disease classes. Bold and italics indicate best and second best respectively for each experimental setting.}
\label{tab:classification}
\end{table}

\subsubsection{Joint Training Encoder and Probe}

After five epochs of supervised adaptation, ViT-B/16 models again outperformed ResNet-101 models. ViT-B/16 VICReg achieved the strongest performance among external-eye-pretrained models, while InfoNCE and LeJEPA remained close. MAE also adapted well, matching the highest accuracy and F1 and achieving the highest recall in the five-epoch setting. In contrast, DINOv2 performed poorly after fine-tuning despite strong frozen-probe performance, suggesting that frozen separability does not necessarily predict short-horizon adaptation behavior. The ResNet-101 models remained below the ViT-B/16 models across objectives. 

Qualitative t-SNE visualizations of fine-tuned embeddings showed disease-labeled structure in the adapted representation space (Figure~\ref{fig:disease_tsne}). MAE produced visually tighter local clusters, although some disease labels appeared split across multiple regions, while VICReg produced a more diffuse embedding with less apparent within-class fragmentation. 

Taken together, these results demonstrate that classification of diseases affecting the external eye is sensitive to both architecture and objective family with external-eye-pretrained ViT-B/16 encoders providing the strongest overall disease-relevant representations compared to baselines.

%% file: merged-sections/7-conclusions.tex
\section{Discussion}

External-eye photographs are increasingly used for clinical classification, anatomical analysis, and oculomics, yet most prior external-eye AI studies remain supervised and task-specific, while ophthalmic foundation-model efforts have largely focused on retinal or device-based imaging \cite{hung2021deep,shu2024artificial,abascal2023development,karlin2023ensemble,moon2022machine,alkhadrawi2024deep,han2025deep,li2022artificial,lee2023differentiating,hui2025deep,zhou2023foundation,chia2024foundation,zhu2025oculomics}. Existing periocular datasets have similarly been developed primarily for iris recognition, periocular biometrics, or specialized acquisition settings rather than for learning general-purpose representations \cite{UBRISV1,UBRISV2,padole2012periocular,zanlorensi2022new,VISOB2,tiong2019periocular,TEYED}. Consequently, most existing systems learn features for a single predefined task and provide limited evidence that those features capture reusable anatomical or clinical information that transfers across datasets, acquisition settings, or downstream endpoints. This limitation is particularly relevant as external-eye imaging is increasingly explored as a low-cost source of clinically informative phenotypes, including biomarkers of systemic health in oculomics. Progress in this setting depends not only on task-specific classifiers, but on representations that capture transferable anatomical and clinically relevant structure across datasets and acquisition domains. We address this gap with a standardized external-eye corpus and benchmark spanning embedding geometry, anatomical landmark recovery, cross-dataset anatomical transfer, and disease classification.

A central contribution of SEEC is the conversion of heterogeneous facial-image
resources into a standardized external-eye pretraining corpus. Although the resulting pretraining corpus is derived primarily from public facial-image datasets rather than prospectively collected clinical photographs, this design provides scale and acquisition diversity that are difficult to obtain from clinical cohorts alone. Importantly, the benchmark explicitly evaluates whether representations learned from these public-source images transfer to clinically acquired external-eye photographs through both the \textit{Clinic} geometry analysis and the independent \textit{Disease} classification task. The substantial changes in representation geometry observed on \textit{Clinic}, together with preserved downstream disease classification performance, emphasize that clinical domain shift is a property to be measured rather than assumed away.

More broadly, our findings show that embedding geometry and downstream utility are related but not interchangeable. Effective rank provides a useful representation-level diagnostic for embedding-space utilization and distribution shift, but anatomical recovery, anatomical transfer, and disease separability require dedicated functional probes. This is reflected by the difference in model rankings across benchmark axes. For example, LeJEPA produced the highest-rank projected embedding spaces, but did not dominate all anatomical or disease-classification probes, while ViT encoders trained with InfoNCE or VICReg achieved stronger performance on specific landmark and classification tasks. Our work also demonstrates that domain-specific external-eye pretraining can encode periocular structure not fully captured by general-purpose visual foundation models. In cross-dataset landmark transfer, the best external-eye-pretrained ViT encoders outperformed DINOv2 and MAE across aggregate and regional landmark metrics despite being trained at substantially smaller scale \cite{oquab2023dinov2,he2022masked}. In disease classification, external-eye-pretrained ViT encoders achieved the strongest frozen-probe performance and remained competitive after short-horizon supervised adaptation. 

Taken together, external-eye photography is low-cost, accessible by any modern smartphone, and commonly obtained during routine clinical documentation \cite{babenko2022detection}. Furthermore, recent work suggests that external-eye photographs may provide clues into systemic biology, which has implications far beyond screening for disease that affects the external eye \cite{babenko2022detection,babenko2023deep}. Realizing this potential will require representations that preserve periocular anatomy, generalize across acquisition settings, and support numerous downstream tasks. The benchmark introduced here provides a foundation for this work by standardizing external-eye pretraining resources, evaluation distributions, anatomical probes, and clinical classification tasks. We hope this resource lowers the barrier to entry for external-eye representation learning and supports broader investigation of the externally visible eye as a scalable modality for medical computer vision.

%% file: merged-sections/6-limitations.tex
\begin{figure}[!t]
    \centering
    \includegraphics[width=.9\linewidth]{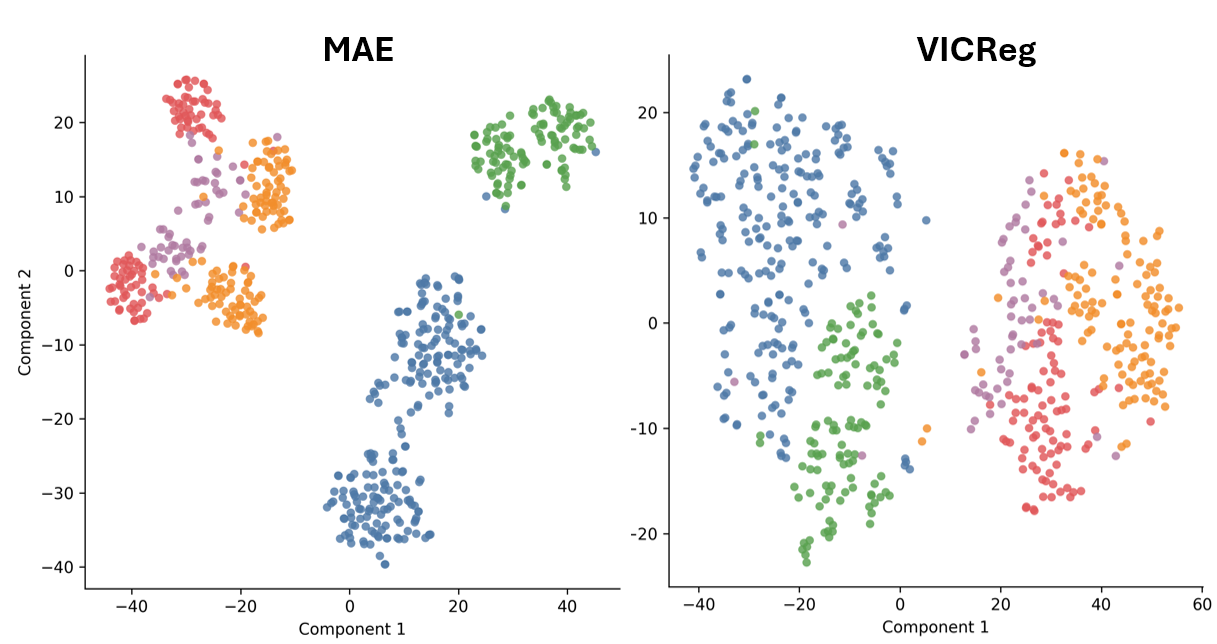}
    \caption{
    Qualitative visualization of disease-labeled embeddings after supervised fine-tuning. t-SNE projections are shown for MAE and VICReg ViT-B encoders after five epochs of fine-tuning on the disease classification task. Points represent external-eye images and are colored by disease class. t-SNE is shown for qualitative visualization only; quantitative comparisons are based on held-out classification metrics in Table~\ref{tab:classification}.
    }
    \label{fig:disease_tsne}
\end{figure}

\section{Limitations and Future Work}

\paragraph{Limitations.}
We have identified several limitations from our work. \textbf{First}, the clinical evaluations are limited in scope. \textit{Disease} contains 633 images across five oculoplastic and craniofacial disease classes and was evaluated using source-image-level grouping rather than confirmed patient-level separation because patient-level identifiers were unavailable. In addition, the clinical domain-shift and disease datasets are not publicly released due to institutional and ethical restrictions, although access may be considered under appropriate institutional review and data-use agreements. More broadly, complete reproduction of all benchmark components is constrained by source-specific licensing and clinical data-use restrictions. Where direct redistribution is not permitted, we provide manifests and reconstruction tools rather than the processed images themselves. \textbf{Second}, the anatomical probe evaluates a defined subset of periocular structure. Landmark targets were deterministically derived from previously annotated segmentation masks and comprise nine iris, scleral, canthal, and eyelid landmarks across two probe datasets. Although this provides a reproducible measure of anatomical recovery, it does not capture the full range of potentially relevant periocular morphology. \textbf{Finally}, qualitative visualization and short-horizon adaptation results should not be overinterpreted. In our t-SNE results (Figure~\ref{fig:disease_tsne}), we observed overlap between some disease classes which suggests that there was not full separation of all classes. Additionally, after five-epoch supervised adaptation, ResNet-101 models lost performance on disease classification, and DINOv2, despite strong frozen-probe performance, degraded substantially under fine-tuning. ResNet-101 degradation likely reflects insufficient capacity to adapt under a small labeled set, while DINOv2's collapse despite strong frozen performance may reflect optimization sensitivity.

\paragraph{Future Work.}

Several extensions follow naturally from our work. Our controlled pretraining experiments use a limited two-view setup to maintain comparability across objective families; future work should evaluate whether richer local/global multi-view training \cite{balestriero2025lejepa}, longer pretraining schedules, and larger models further improve performance. In addition, the disease classification probe is intentionally standardized rather than optimized for maximum supervised performance. Future work should evaluate stronger adaptation protocols using larger labeled datasets, and investigate the different adaptation behavior observed across architectures. 


%% file: merged-sections/99-release-and-maintenance.tex
\section{Release, Licensing, and Maintenance}

\subsection{Licensing and responsibility}

The SEEC dataset and benchmark toolkits are released under the MIT License. Original source images and reconstructed periocular crops remain subject to the terms of their respective source datasets. The released reconstruction archive contains processing records required to reproduce SEEC from permitted source data; it does not redistribute source images. The released checkpoint archive contains SEEC-trained model weights. The authors accept responsibility for the materials they distribute and for compliance with applicable rights and licenses.

\subsection{Maintenance}

We intend to maintain the repositories, documentation, and hosted release artifacts, including correcting broken links, documentation errors, and reproducibility issues. Issues may be reported through the public GitHub repositories or to the corresponding authors.

%% file: merged-sections/99-datasheet.tex
\section{Dataset Documentation}

\subsection{Purpose and composition}

SEEC (Standardized External-Eye Corpus) is a large-scale external-eye/periocular image corpus developed for representation learning and benchmarking tasks related to the external eye. The corpus contains 4,352,692 unilateral 224$\times$224 periocular crops derived from 14 public face-image datasets, together with a 134,969-image high-resolution subset at 512$\times$512. Counts refer to images rather than unique individuals.

The accompanying benchmark includes the following resources:

\begin{itemize}
    \item Pretrain-10K: 10,000 images
    \item Pretrain-100K: 100,000 images
    \item Pretrain-1M: 1,000,000 images
    \item Holdout: 421,730 images
    \item Open-HR: 134,969 images
    \item Clinic: 35,082 restricted clinical images
    \item LM-Celeb: 3,610 landmark-annotated images
    \item LM-CFD: 1,596 landmark-annotated images
    \item Disease: 633 restricted clinical images across five classes
\end{itemize}

These resources serve different experimental roles and should not be summed as independent additions to the corpus.

\subsection{Source datasets and construction}

SEEC was constructed from CelebA, UTKFace, FFHQ, VGGFace2, Tufts Face Database, Chicago Face Database, CFD-MR, CFD-India, IMDB, WIKI, UMDFaces, Face Images with Marked Landmark Points, FEI Face Database, and MORPH.

Corpus construction included deduplication, face detection and alignment, manual periocular quality annotation, dataset-specific CNN filtering, bilateral periocular cropping, left/right splitting, and resizing. Released reconstruction manifests record the transformations and dataset membership required to reconstruct the public portions of SEEC from source data. Source images are not redistributed by SEEC.

\subsection{Benchmark use}

The SEEC benchmark evaluates learned representations across four settings: embedding geometry under distribution shift, anatomical landmark recovery, cross-dataset anatomical transfer, and disease classification using frozen probes and short supervised fine-tuning.

Public benchmark resources are intended to support representation-learning research, model comparison, and reproducibility studies involving external-eye and periocular images.

\subsection{Access and documentation}

The SEEC project page is available at \url{https://monkeygobah.github.io/seec-project/}. The dataset reconstruction toolkit is maintained at \url{https://github.com/monkeygobah/seec-dataset}, and the benchmark code and fixed evaluation resources are maintained at \url{https://github.com/monkeygobah/seec-benchmark}. Reconstruction manifests and trained checkpoints are hosted on Google Drive and are linked from the project page. The restricted clinical resources used for the Clinic and Disease evaluations are not publicly distributed.

\subsection{Intended use, limitations, and human data}

SEEC is intended to support research on external-eye and periocular representation learning and benchmarking. Its primary use is as a standardized substrate for studying whether models trained on large-scale external-eye imagery learn anatomical and clinically relevant representations that transfer across datasets and tasks.

SEEC is derived from previously released face-image datasets rather than from direct collection of new public-subject data by the SEEC authors. Collection procedures, consent mechanisms, demographic composition, and original terms therefore differ across source datasets and should be interpreted according to the documentation provided by the respective dataset creators.

Because SEEC combines images from heterogeneous public face datasets, its scale should not be interpreted as evidence of demographic or clinical representativeness. Periocular cropping removes much of the surrounding facial context but does not guarantee anonymity or eliminate the possibility of identification. SEEC does not provide a harmonized demographic characterization or validated subject-level metadata across the combined corpus.

The accompanying clinical evaluations are intended to assess representation quality rather than establish deployment-ready clinical performance. The Clinic and Disease datasets are restricted and are not publicly distributed. SEEC is not intended for face recognition, identity verification, or direct clinical deployment.

%% file: merged-sections/99-supp-methods.tex
\section{Supplemental Methods}

\subsection{Further Details on SSL Implementation}
\label{app:ssl_objectives}

All objectives are trained on the same standardized external-eye crops under the same fixed-compute training budget described in Section~\ref{model-setup}. Unless otherwise noted, all methods use the same ResNet-101 or ViT backbone, the same three-layer projection head with hidden dimension 2048 and output dimension 128, and the same optimizer and learning-rate schedule.

\subsubsection{Shared View Generation}

For all self-supervised objectives, each 224$\times$224 unilateral external-eye crop is transformed into two local views using random resized cropping with output size 112$\times$112 and scale range $[0.08, 1.0]$, followed by ImageNet normalization. These paired views are used as the two augmented observations of the same underlying image for self-supervised training.

\subsubsection{InfoNCE}

The contrastive baseline uses the InfoNCE objective. Let $z_i^{(1)}$ and $z_i^{(2)}$ denote the projected embeddings of the two augmented views of image $i$ in a minibatch of size $B$. 

We use cosine similarity in the projected embedding space and optimize the symmetric two-view InfoNCE loss
\begin{equation}
\mathcal{L}_{\mathrm{InfoNCE}}
=
\frac{1}{2B}
\sum_{i=1}^{B}
\left[
-\log
\frac{\exp(\mathrm{sim}(z_i^{(1)}, z_i^{(2)})/\tau)}
{\sum_{j=1}^{B}\exp(\mathrm{sim}(z_i^{(1)}, z_j^{(2)})/\tau)}
-\log
\frac{\exp(\mathrm{sim}(z_i^{(2)}, z_i^{(1)})/\tau)}
{\sum_{j=1}^{B}\exp(\mathrm{sim}(z_i^{(2)}, z_j^{(1)})/\tau)}
\right],
\end{equation}
where $\mathrm{sim}(\cdot,\cdot)$ denotes cosine similarity and $\tau$ is a temperature parameter. In all experiments, we set $\tau = 0.2$.

\subsubsection{VICReg}

The redundancy-reduction baseline uses VICReg, which combines invariance, variance, and covariance regularization terms. Let $z^{(1)}\in \mathbb{R}^{N \times D}$ and $z^{(2)}\in \mathbb{R}^{N \times D}$ denote the projected embeddings from the two augmented views in a minibatch. The VICReg loss is
\begin{equation}
\mathcal{L}_{\mathrm{VICReg}}
=
\lambda \mathcal{L}_{\mathrm{inv}}
+
\mu \mathcal{L}_{\mathrm{var}}
+
\nu \mathcal{L}_{\mathrm{cov}},
\end{equation}
where the invariance term is the mean squared difference between paired embeddings,
\begin{equation}
\mathcal{L}_{\mathrm{inv}}
=
\frac{1}{B}\sum_{i=1}^{B}
\left\| z_i^{(1)} - z_i^{(2)} \right\|_2^2,
\end{equation}
where $z_i^{(v)}\in\mathbb{R}^D$ is the $i-$th row of view $z^{(v)}, v = 1,2.$
the variance term is computed independently on each view and penalizes dimensions whose batch standard deviation falls below a target threshold,
\begin{equation}
\mathcal{L}_{\mathrm{var}}
=
\frac{1}{d}\sum_{k=1}^{d}
\left[
\max(0, \gamma - \sigma_k(z^{(1)}))
+
\max(0, \gamma - \sigma_k(z^{(2)}))
\right],
\end{equation}
where $d$ is the projection dimension, $\sigma_k(\cdot)$ is the batch standard deviation of dimension $k$, and $\gamma$ is the target standard deviation.


And the covariance term penalizes off-diagonal entries of the empirical covariance matrix of each view,
\begin{equation}
\mathcal{L}_{\mathrm{cov}}
=
\frac{1}{d}\sum_{a \neq b}
\left[
\mathrm{Cov}(z^{(1)})_{ab}^2
+
\mathrm{Cov}(z^{(2)})_{ab}^2
\right],
\end{equation}
where $\mathrm{Cov}(z^v)=z^{(v)\top}z^{(v)}\in \mathbb{R}^{D \times D}, v =1, 2$.

In all experiments, we use $\lambda = 25$, $\mu = 25$, $\nu = 1$, target standard deviation $\gamma = 1.0$, and $\epsilon = 10^{-4}$.

\subsubsection{LeJEPA}
The predictive baseline uses LeJEPA, a JEPA-style objective operating on paired augmented views of the same image \cite{balestriero2025lejepa}. Given two views, the encoder and projector produce view-specific embeddings, and training encourages these embeddings to agree while preventing representational collapse. In the benchmark configuration used in this paper, the collapse-prevention term is the sliced Gaussian regularizer SigReg \cite{balestriero2025lejepa}.

Let $\mathcal{L}_{\mathrm{sim}}$ denote the predictive similarity term between paired projected views, and let $\mathcal{L}_{\mathrm{SigReg}}$ denote the SigReg regularizer. The overall LeJEPA objective is
\begin{equation}
\mathcal{L}_{\mathrm{LeJEPA}}
=
(1-\lambda)\mathcal{L}_{\mathrm{sim}}
+
\lambda \mathcal{L}_{\mathrm{SigReg}},
\end{equation}
where $\lambda$ controls the tradeoff between predictive agreement and regularization. In all experiments, we set $\lambda = 0.05$. The similarity term is implemented as the mean squared deviation of each projected view embedding from the per-image average projected embedding across views. We follow default implementation for SigReg and use 17 spline knots and 256 sampled slices. 

In our framework, to be compared directly with InfoNCE and VicReg LeJEPA is trained using two augmented views per image to match the paired-view setup used for the other self-supervised baselines.

\subsubsection{Shared Optimization Settings}

All objective families are trained with AdamW using learning rate $5\times10^{-4}$ and weight decay $5\times10^{-4}$. We use a 12{,}500-step linear warmup with start factor 0.01, followed by cosine learning-rate decay to a final learning rate of $5\times10^{-7}$. All experiments use a global batch size of 384 and are trained for 50{,}000 optimization steps.

\subsection{Further Details on Embedding Geometry Metrics}
\label{app:embedding_geometry}

For each frozen model and evaluation distribution, we extracted projected embeddings $X \in \mathbb{R}^{N \times D}$ and computed geometry metrics on the centered embeddings $\tilde{X}=X-\bar{X}$. Let $\lambda_1 \geq \cdots \geq \lambda_D$ denote the eigenvalues of the embedding covariance matrix $C$ of $C=\tilde{X}^\top \tilde{X}$.

We summarize covariance-spectrum usage using the participation-ratio effective rank,
\[
r_{\mathrm{eff}}=\frac{\left(\sum_{i=1}^{D}\lambda_i\right)^2}{\sum_{i=1}^{D}\lambda_i^2},
\]
and report $r_{\mathrm{eff}}/D$ in the main text. Higher values indicate that variance is spread across more embedding dimensions, whereas lower values indicate concentration of variance in fewer directions. We also compute explained-variance concentration in the leading principal components,
\[
\mathrm{EV@}k=\frac{\sum_{i=1}^{k}\lambda_i}{\sum_{i=1}^{D}\lambda_i},
\]
for $k \in \{1,5,20\}$.

To characterize angular structure, we L2-normalize centered embeddings and sample up to 200{,}000 random pairs per model and dataset using a fixed seed. From these pairs, we compute the mean and standard deviation of cosine similarity.

\subsection{Anatomical Landmark Probe Details}
\label{app:landmark_probe}

For each frozen encoder, embeddings were extracted from \textit{LM-Celeb} and \textit{LM-CFD} using the fixed image preprocessing pipeline described in Section~\ref{sec:probes}. The encoder weights were held fixed for all landmark experiments. For ResNet-101 models, features were extracted from the frozen backbone feature map. We evaluated global average pooling and $4\times4$ adaptive grid pooling. For ViT models, patch tokens were reshaped into spatial feature maps and pooled using the same $4\times4$ grid-pooling procedure.

The landmark probe was implemented as a multilayer perceptron with hidden dimensions 1024 and 512 and dropout 0.2. The probe predicted 18 output values corresponding to $(x,y)$ coordinates for nine landmarks: iris centroid, iris superior, iris inferior, sclera superior, sclera inferior, medial canthus, lateral canthus, lid superior, and lid inferior. Probes were trained with AdamW using learning rate $10^{-3}$ and weight decay $10^{-4}$ for up to 1000 epochs, with early stopping based on validation mean landmark Euclidean error.

We report mean absolute error and mean Euclidean error over landmarks. For each sample and landmark, Euclidean error was computed in pixel units on the 224$\times$224 crop coordinate system. Within-dataset evaluation used fixed train, validation, and test splits from the same source dataset. Cross-dataset transfer used the \textit{LM-Celeb} training and validation splits for probe fitting and model selection, followed by evaluation on the \textit{LM-CFD} test split without retraining or further adaptation.

%% file: merged-sections/99-supp-results.tex
\section{Supplementary Results}

\subsection{SEEC Construction}

\begin{table}[!t]
\centering
\begin{tabular}{@{}cccccc@{}}
\toprule
\textbf{Dataset} &
\makecell{\textbf{Source}\\\textbf{Images}} &
\makecell{\textbf{Exact}\\\textbf{Duplicates}} &
\makecell{\textbf{Images After}\\\textbf{Duplicate Removal}} &
\makecell{\textbf{Grayscale}\\\textbf{(\%)}} &
\makecell{\textbf{RGB}\\\textbf{(\%)}} \\
\midrule

\textbf{CFD}        & 1,207     & 0       & 1,207     & 0.00   & 100.00 \\
\textbf{CFD-I}      & 146       & 0       & 146       & 0.00   & 100.00 \\
\textbf{CFD-MR}     & 88        & 0       & 88        & 0.00   & 100.00 \\
\textbf{Tufts}      & 560       & 0       & 560       & 0.00   & 100.00 \\
\textbf{FEI}        & 2,800     & 0       & 2,800     & 0.00   & 100.00 \\
\textbf{FIML}       & 7,049     & 555     & 6,494     & 100.00 & 0.00   \\
\textbf{UTKFace}    & 24,106    & 329     & 23,777    & 0.05   & 99.95  \\
\textbf{MORPH}      & 57,298    & 227     & 57,071    & 2.95   & 97.05  \\
\textbf{WikiFace}   & 62,328    & 9,746   & 52,582    & 20.85  & 79.15  \\
\textbf{CelebA}     & 202,599   & 131     & 202,468   & 0.00   & 100.00 \\
\textbf{UMD Faces}  & 367,888   & 0       & 367,888   & 0.00   & 100.00 \\
\textbf{IMDB-WIKI}  & 460,723   & 155,311 & 305,412   & 4.89   & 95.11  \\
\textbf{FFHQ}       & 70,000    & 0       & 70,000    & 0.00   & 100.00 \\
\textbf{VGGFace2}   & 3,137,807 & 3,084   & 3,134,723 & 0.00   & 100.00 \\
\bottomrule
\end{tabular}%
\caption{Source dataset composition following duplicate removal. The table reports the number of images contributed by each source dataset, the number of exact duplicates identified within each source dataset, the number of images retained after duplicate removal, and image-format distributions.}
\label{tab:supp-source-composition}
\end{table}

\begin{table}[!t]
\centering
\resizebox{\columnwidth}{!}{%
\begin{tabular}{@{}cccccc@{}}
\toprule
\textbf{Dataset} &
\makecell{\textbf{Initial Positives}\\\textbf{(Grader 1)}} &
\makecell{\textbf{Initial Negatives}\\\textbf{(Grader 1)}} &
\makecell{\textbf{Positive Agreement}\\\textbf{($n=100$)}} &
\makecell{\textbf{Negative Agreement}\\\textbf{($n=100$)}} &
\makecell{\textbf{Negative Regrade Agreement}\\\textbf{(All Negatives)}} \\
\midrule

\textbf{CelebA}    & 1,449 & 1,051 & 0.99 & 0.57 & 0.47 \\
\textbf{FIML}      & 940   & 1,560 & 0.97 & 0.68 & 0.66 \\
\textbf{IMDB-WIKI} & 952   & 1,548 & 1.00 & 0.59 & 0.55 \\
\textbf{MORPH}     & 2,399 & 101   & 0.99 & 0.53 & 0.28 \\
\textbf{UMD Faces} & 1,309 & 1,191 & 0.99 & 0.59 & 0.44 \\
\textbf{UTKFace}   & 1,808 & 692   & 1.00 & 0.39 & 0.33 \\
\textbf{WikiFace}  & 990   & 1,510 & 0.98 & 0.64 & 0.62 \\
\textbf{FFHQ}      & 2,305 & 195   & 0.95 & 0.95 & 0.95 \\
\textbf{VGGFace2}  & 1,946 & 554   & 0.96 & 0.91 & 0.95 \\
\bottomrule
\end{tabular}%
}
\caption{Manual periocular annotation and intergrader agreement by source dataset. Grader 1 labeled 2,500 randomly sampled images per dataset as periocular or non-periocular. A second blinded grader independently reassessed randomly sampled positive and negative images and all images initially classified as negative.}
\label{tab:supp-grader-agreement}
\end{table}

\begin{table}[!t]
\centering
\resizebox{.4\columnwidth}{!}{%
\begin{tabular}{@{}cccc@{}}
\toprule
\textbf{Dataset} & \textbf{Positive} & \textbf{Negative} & \textbf{Gray} \\
\midrule
\textbf{CelebA}    & 1,449 & 494   & 557 \\
\textbf{FIML}      & 940   & 1,028 & 532 \\
\textbf{IMDB-WIKI} & 952   & 845   & 703 \\
\textbf{MORPH}     & 2,399 & 28    & 73  \\
\textbf{UMD Faces} & 1,309 & 523   & 668 \\
\textbf{UTKFace}   & 1,808 & 228   & 464 \\
\textbf{WikiFace}  & 990   & 932   & 578 \\
\textbf{FFHQ}      & 2,305 & 186   & 9   \\
\textbf{VGGFace2}  & 1,946 & 527   & 27  \\
\bottomrule
\end{tabular}%
}
\caption{Final periocular annotation distribution following blinded regrading. Images were categorized as periocular, non-periocular, or ambiguous (gray).}
\label{tab:supp-final-labels}
\end{table}

\begin{table}[!t]
\centering
\resizebox{\columnwidth}{!}{%
\begin{tabular}{@{}ccccccccc@{}}
\toprule
\textbf{Dataset} &
\textbf{Accuracy} &
\textbf{Precision} &
\textbf{Recall} &
\textbf{F1} &
\makecell{\textbf{AUROC}\\\textbf{(95\% CI)}} &
\makecell{\textbf{Selected}\\\textbf{Threshold}} &
\makecell{\textbf{Selected}\\\textbf{FPR}} &
\makecell{\textbf{Selected}\\\textbf{TPR}} \\
\midrule

\textbf{CelebA}    & 0.95 & 0.97 & 0.93 & 0.95 & 0.99 [0.98--1.00] & 0.92 & 0.014 & 0.85 \\
\textbf{FIML}      & 0.97 & 0.96 & 0.97 & 0.97 & 0.99 [0.98--1.00] & 0.95 & 0.028 & 0.93 \\
\textbf{IMDB-WIKI} & 0.95 & 0.96 & 0.94 & 0.95 & 0.99 [0.98--1.00] & 0.70 & 0.024 & 0.92 \\
\textbf{MORPH}     & 1.00 & 1.00 & 1.00 & 1.00 & 1.00 [1.00--1.00] & 0.50 & 0.000 & 1.00 \\
\textbf{UMD Faces} & 0.96 & 0.95 & 0.97 & 0.96 & 1.00 [0.99--1.00] & 0.74 & 0.013 & 0.97 \\
\textbf{UTKFace}   & 0.85 & 0.80 & 0.94 & 0.86 & 0.95 [0.88--0.99] & 0.89 & 0.000 & 0.59 \\
\textbf{WikiFace}  & 0.95 & 0.95 & 0.96 & 0.95 & 0.99 [0.99--1.00] & 0.72 & 0.021 & 0.94 \\
\textbf{FFHQ}      & 0.88 & 0.86 & 0.89 & 0.88 & 0.97 [0.92--1.00] & 0.95 & 0.036 & 0.79 \\
\textbf{VGGFace2}  & 0.89 & 0.92 & 0.86 & 0.89 & 0.96 [0.93--0.98] & 0.95 & 0.013 & 0.71 \\
\bottomrule
\end{tabular}%
}
\caption{Performance and selected operating thresholds of dataset-specific ResNet-18 classifiers for periocular image detection. Accuracy, precision, recall, F1 score, and area under the receiver operating characteristic curve (AUROC) with 95\% confidence intervals are reported on the held-out test sets. FPR and TPR denote false-positive and true-positive rates, respectively, at the selected operating threshold.}
\label{tab:supp-cnn-perf-thresholds}
\end{table}

\begin{figure}[!t]
    \centering
    \includegraphics[width=\linewidth]{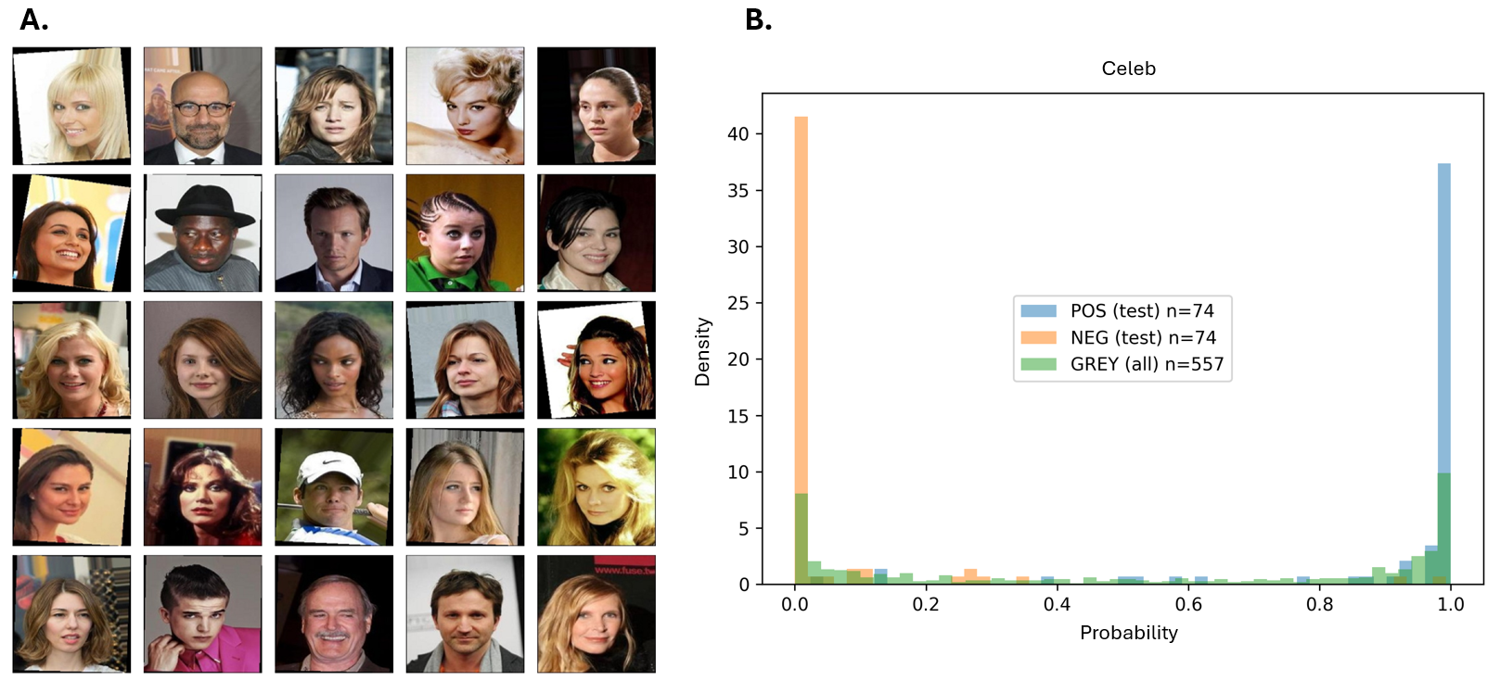}
    \caption{Qualitative assessment of CNN-based periocular filtering. 
    (A) Representative low-confidence images retained by the dataset-specific classifiers near the selected probability threshold. 
    (B) Example distributions of predicted periocular probabilities for held-out positive and negative samples and manually designated ambiguous images. 
    These analyses were used to assess whether the selected operating thresholds limited retention of non-periocular images while preserving images with discernible periocular anatomy.}
    \label{fig:supp-filtering-qc}
\end{figure}

\begin{table}[!t]
\centering
\resizebox{\columnwidth}{!}{%
\begin{tabular}{@{}ccccccc@{}}
\toprule
\textbf{Dataset} &
\makecell{\textbf{Original}\\\textbf{Images}} &
\makecell{\textbf{Post-}\\\textbf{Deduplication}} &
\makecell{\textbf{Post-Alignment/}\\\textbf{Preprocessing}} &
\makecell{\textbf{Post-Periocular}\\\textbf{Filtering}} &
\makecell{\textbf{224$\times$224}\\\textbf{Unilateral}} &
\makecell{\textbf{512$\times$512}\\\textbf{High Resolution}} \\ 
\midrule

\textbf{CFD}        & 1,207     & 1,207     & 1,207     & 1,207     & 2,414     & 2,414   \\
\textbf{CFD-I}      & 146       & 146       & 146       & 146       & 292       & 292     \\
\textbf{CFD-MR}     & 88        & 88        & 88        & 88        & 176       & 176     \\
\textbf{Tufts}      & 560       & 560       & 336       & 336       & 672       & 672     \\
\textbf{FEI}        & 2,800     & 2,800     & 1,182     & 1,182     & 2,364     & 4       \\
\textbf{FIML}       & 7,049     & 6,494     & 6,317     & 2,870     & 5,740     & 5,732   \\
\textbf{UTKFace}    & 24,106    & 23,777    & 23,547    & 12,186    & 24,372    & 8,724   \\
\textbf{MORPH}      & 57,298    & 57,071    & 56,974    & 50,627    & 101,254   & 13,759  \\
\textbf{WikiFace}   & 62,328    & 52,582    & 43,329    & 20,514    & 41,028    & 116     \\
\textbf{CelebA}     & 202,599   & 202,468   & 201,932   & 117,844   & 235,688   & 0       \\
\textbf{UMD Faces}  & 367,888   & 367,888   & 361,581   & 212,922   & 425,844   & 4       \\
\textbf{IMDB-WIKI}  & 460,723   & 305,412   & 259,052   & 131,470   & 262,940   & 3,360   \\
\textbf{FFHQ}       & 70,000    & 70,000    & 69,988    & 49,897    & 99,794    & 99,716  \\
\textbf{VGGFace2}   & 3,137,807 & 3,134,723 & 3,065,030 & 1,575,057 & 3,150,114 & 0       \\ 
\midrule
\textbf{Total}      & 4,394,599 & 4,225,216 & 4,090,709 & 2,176,346 & 4,352,692 & 134,969 \\ 
\bottomrule
\end{tabular}%
}
\caption{Source-level image counts across the principal stages of SEEC construction. Counts are reported for the original source collections, after exact duplicate removal, after face detection, alignment, and source-specific preprocessing, after periocular filtering, and after generation of the standardized unilateral $224\times224$ and high-resolution $512\times512$ corpora.}
\label{tab:supp-final-composition}
\end{table}

\begin{table}[]
\centering
\resizebox{\textwidth}{!}{%
\begin{tabular}{@{}ccc@{}}
\toprule
\textbf{Dataset}            & \textbf{Source use terms}                   & \textbf{Redistribution provision} \\ \midrule
CelebA                      & Non-commercial research use                 & Restricted                        \\
UTKFace                     & Non-commercial research use                 & Not expressly specified           \\
FFHQ & Mixed Creative Commons/source-image licenses & Permitted subject to applicable license terms \\
VGGFace2                    & Research use                                & Not expressly specified           \\
Tufts Face Database         & Non-commercial research and educational use & Restricted                        \\
Chicago Face Database (CFD) & Non-commercial scientific research use      & Restricted                        \\
CFD-MR                      & Non-commercial scientific research use      & Restricted                        \\
CFD-I                  & Non-commercial scientific research use      & Restricted                        \\
IMDB                        & Academic research use                       & Not expressly specified           \\
WIKI                        & Academic research use                       & Not expressly specified           \\
UMD Faces                   & Non-commercial research use                 & Restricted                        \\
FIML                        & Research/competition data terms             & Not expressly specified           \\
FEI Face Database           & Research use                                & Restricted                        \\
MORPH (Academic)            & Non-commercial academic use                 & Not expressly specified           \\ \bottomrule
\end{tabular}%
}
\caption{Source-specific use and redistribution terms for datasets contributing to SEEC. ``Restricted'' indicates that redistribution is expressly limited or prohibited by the source terms; ``Not expressly specified'' indicates that research use is permitted but redistribution of derived images is not explicitly addressed.}
\label{tab:supp-license}
\end{table}

\subsection{Additional Embedding Geometry Metrics}
\label{app:embedding_geometry_results}

\begin{figure}[!t]
    \centering
    \includegraphics[width=1\linewidth]{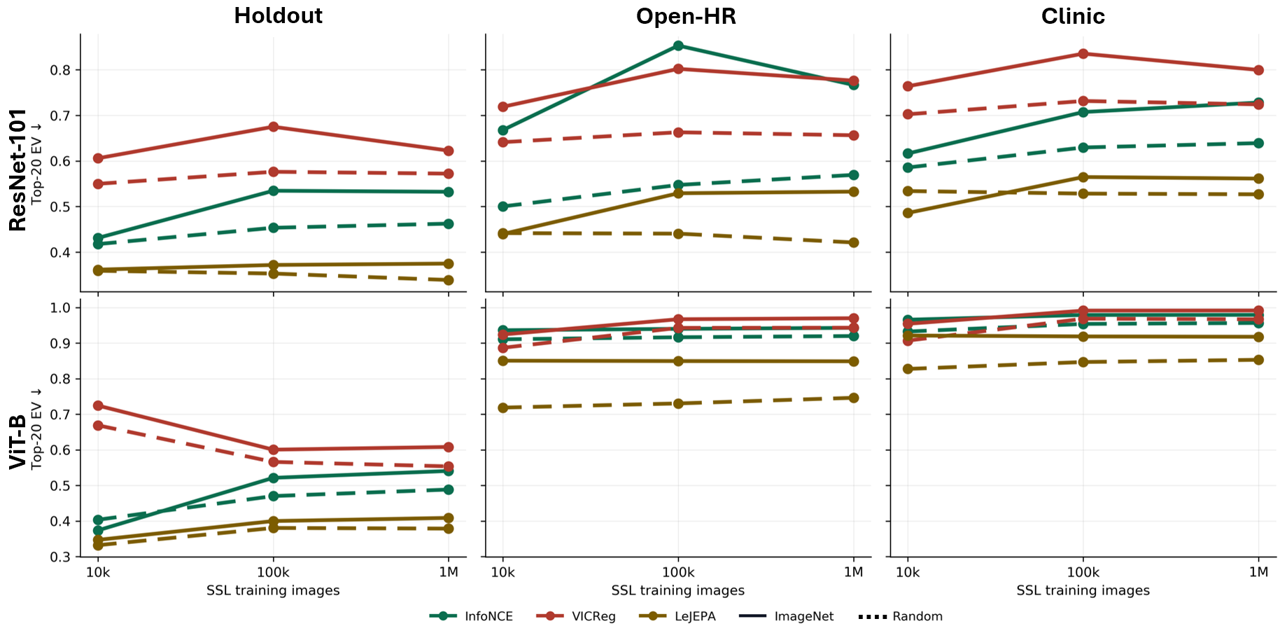}
    \caption{Top-20 explained variance of projected embeddings across architecture, objective, initialization, pretraining scale, and evaluation distribution. Lower values indicate less concentration of variance in the leading covariance directions.}
    \label{fig:geometry-ev20-supp}
\end{figure}

\begin{figure}[!t]
    \centering
    \includegraphics[width=1\linewidth]{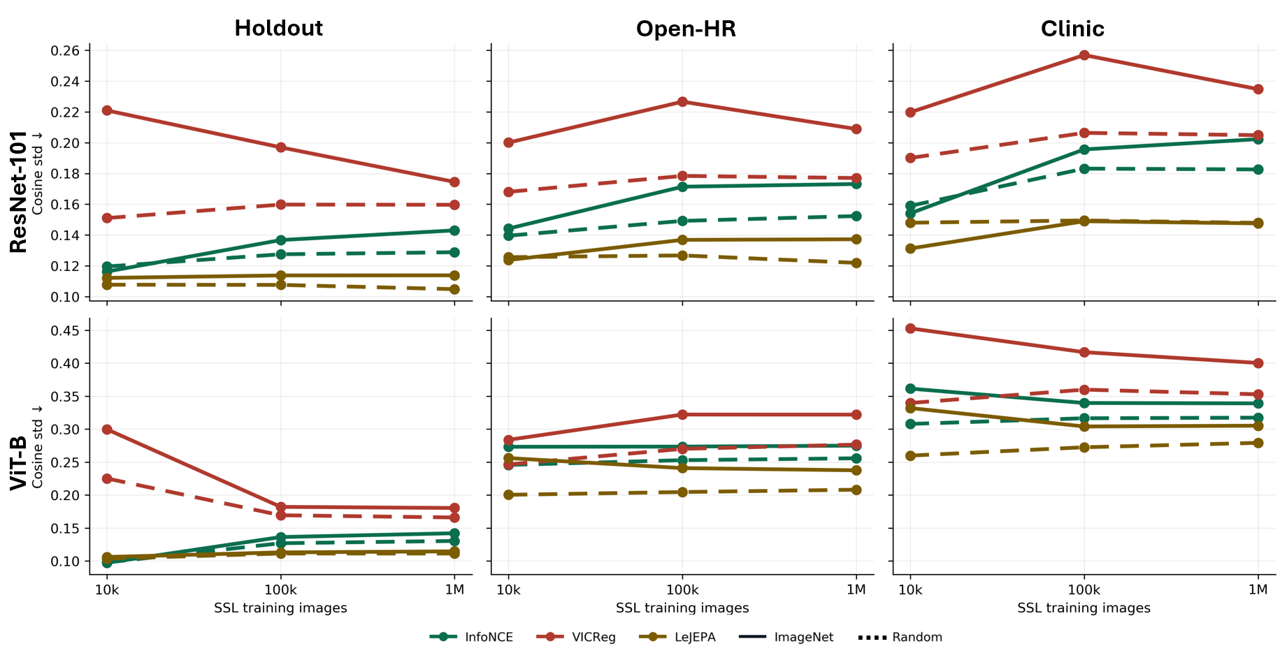}
    \caption{Standard deviation of sampled pairwise cosine similarities after centering and row-normalizing projected embeddings. Lower values indicate less angular concentration among samples.}
    \label{fig:geometry-cos-supp}
\end{figure}

In addition to normalized effective rank, we evaluated complementary geometry diagnostics based on spectral concentration and pairwise angular dispersion (Figure~\ref{fig:geometry-ev20-supp} and ~\ref{fig:geometry-cos-supp}). These analyses support the same qualitative conclusions as the main-text effective-rank results. Across architectures and evaluation distributions, LeJEPA generally produced lower top-20 explained variance, indicating that variance was less concentrated in the leading covariance directions. Conversely, VICReg and some InfoNCE configurations showed higher top-20 explained variance, consistent with more anisotropic projected embedding spaces.

Pairwise cosine dispersion showed a similar pattern. After centering and row-normalizing projected embeddings, LeJEPA generally produced lower cosine standard deviation than InfoNCE or VICReg, indicating less angular concentration among samples. The separation between objectives was most apparent under distribution shift, particularly on \textit{Clinic}, where several configurations exhibited increased spectral and angular concentration. 

\begin{figure}[!t]
    \centering
    \includegraphics[width=1\linewidth]{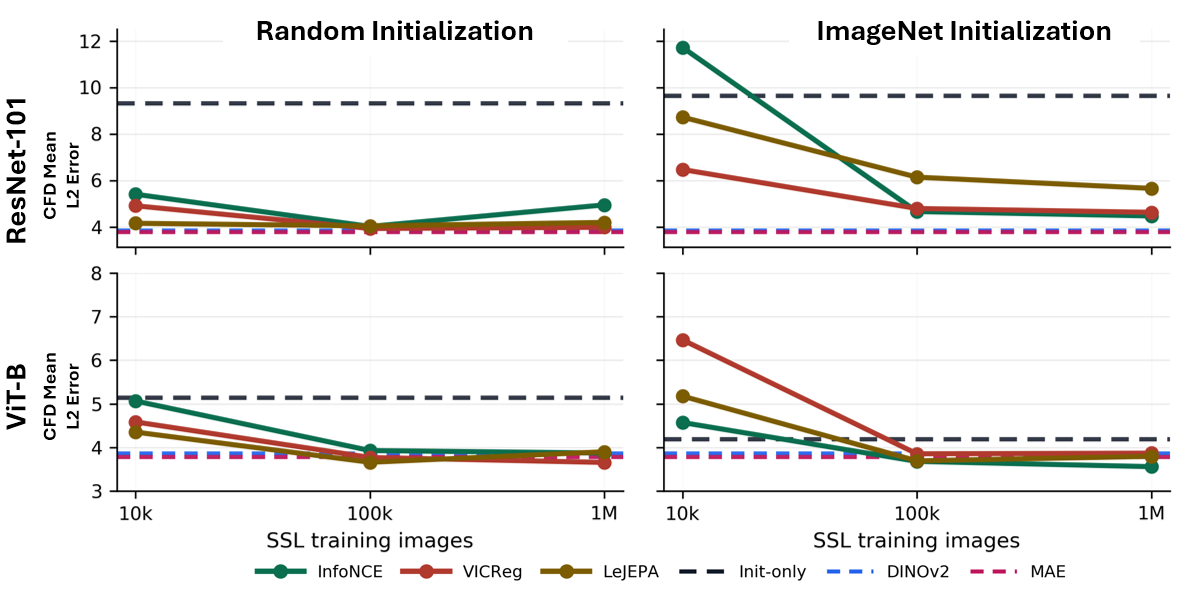}
    \caption{Within-dataset LM-CFD landmark probe performance for ResNet-101 and ViT-B/16 encoders. ViT models achieve strong landmark prediction performance, with external DINOv2 and MAE baselines shown as dashed references.}
    \label{fig:cfd-within}
\end{figure}